\documentclass[manuscript,screen,nonacm]{acmart}%

\AtBeginDocument{%
  }

\setcopyright{rightsretained}
\acmJournal{JRC}
\acmYear{2024} \acmVolume{1} \acmNumber{1} \acmArticle{1} \acmMonth{1}\acmDOI{10.1145/3677173}

\newcommand\coderepo{\url{https://github.com/So-Cool/CrossModelFairness}}

\usepackage{todonotes}

\usepackage[inline]{enumitem}

\usepackage{booktabs}
\usepackage{multirow}

\usepackage{wrapfig}

\usepackage{caption}
\usepackage{subcaption}
\usepackage{dblfloatfix}
\usepackage{placeins}

\usepackage{hyperref}

\usepackage{mathtools}
\usepackage{dsfont}

\newtheorem{definition}{Definition}

\newtheorem{proposition}{Proposition}

\DeclareMathOperator*{\argmax}{arg\,max}

\graphicspath{{./figure/}}

\usepackage[framemethod=tikz]{mdframed}
\usepackage{fontawesome}

\definecolor{mycolor}{rgb}{0.122, 0.435, 0.698}
\definecolor{myblue}{rgb}{0.122, 0.435, 0.698}
\definecolor{mygreen}{rgb}{0.125, 0.525, 0.220}
\definecolor{myyellow}{rgb}{0.588, 0.439, 0.000}
\definecolor{myred}{rgb}{0.647, 0.114, 0.165}
\definecolor{myviolet}{rgb}{0.71764706, 0.40784314, 0.63529412}

\newmdenv[innerlinewidth=0.5pt,roundcorner=4pt,innerleftmargin=4.25pt,
          innerrightmargin=4.25pt,innertopmargin=4.25pt,innerbottommargin=4.25pt,
          linecolor=mycolor,backgroundcolor=mycolor!25!white]{mybox}
\newmdenv[innerlinewidth=0.5pt,roundcorner=4pt,innerleftmargin=4.25pt,
          innerrightmargin=4.25pt,innertopmargin=4.25pt,innerbottommargin=4.25pt,
          linecolor=myblue,backgroundcolor=myblue!25!white]{mybluebox}
\newmdenv[innerlinewidth=0.5pt,roundcorner=4pt,innerleftmargin=4.25pt,
          innerrightmargin=4.25pt,innertopmargin=4.25pt,innerbottommargin=4.25pt,
          linecolor=mygreen,backgroundcolor=mygreen!25!white]{mygreenbox}
\newmdenv[innerlinewidth=0.5pt,roundcorner=4pt,innerleftmargin=4.25pt,
          innerrightmargin=4.25pt,innertopmargin=4.25pt,innerbottommargin=4.25pt,
          linecolor=myyellow,backgroundcolor=myyellow!25!white]{myyellowbox}
\newmdenv[innerlinewidth=0.5pt,roundcorner=4pt,innerleftmargin=4.25pt,
          innerrightmargin=4.25pt,innertopmargin=4.25pt,innerbottommargin=4.25pt,
          linecolor=myred,backgroundcolor=myred!25!white]{myredbox}
\newmdenv[innerlinewidth=0.5pt,roundcorner=4pt,innerleftmargin=4.25pt,
          innerrightmargin=4.25pt,innertopmargin=4.25pt,innerbottommargin=4.25pt,
          linecolor=myviolet,backgroundcolor=myviolet!25!white]{myvioletbox}

\newlength\stextwidth
\newcommand\makesamewidth[3][c]{%
  \settowidth{\stextwidth}{#2}%
  \makebox[\stextwidth][#1]{#3}%
}

\begin{document}

\title[%
Cross-model Fairness: Empirical Study of Fairness and Ethics Under Model Multiplicity%
]{%
Cross-model Fairness:\\ Empirical Study of Fairness and Ethics Under Model Multiplicity%
}

\author{Kacper Sokol}
\email{kacper.sokol@inf.ethz.ch}
\orcid{0000-0002-9869-5896}
\affiliation{%
  \institution{ARC Centre of Excellence for Automated Decision-Making and Society, RMIT University}
  \city{Melbourne}
  \country{Australia}
}
\affiliation{%
  \institution{Department of Computer Science, ETH Zurich}
  \country{Switzerland}
}

\author{Meelis Kull}
\email{meelis.kull@ut.ee}
\orcid{0000-0001-9257-595X}
\affiliation{%
  \institution{Institute of Computer Science, University of Tartu}
  \country{Estonia}
}

\author{Jeffrey Chan}
\email{jeffrey.chan@rmit.edu.au}
\orcid{0000-0002-7865-072X}
\affiliation{%
  \institution{ARC Centre of Excellence for Automated Decision-Making and Society, RMIT University}
  \city{Melbourne}
  \country{Australia}
}

\author{Flora Salim}%
\email{flora.salim@unsw.edu.au}
\orcid{0000-0002-1237-1664}
\affiliation{%
  \institution{ARC Centre of Excellence for Automated Decision-Making and Society, UNSW Sydney}%
  \country{Australia}
}

\renewcommand{\shortauthors}{Kacper Sokol, Meelis Kull, et al.}

\begin{abstract}
While data-driven predictive models are a strictly technological construct, they may operate within a social context in which benign engineering choices entail implicit, indirect and unexpected real-life consequences. %
Fairness of such systems -- pertaining both to individuals and groups -- is one relevant consideration in this space; %
algorithms can discriminate people across various \emph{protected} characteristics regardless of whether these properties are included in the data or discernible through proxy variables. %
To date, this notion has predominantly been studied for a \emph{fixed} model, often under different classification thresholds, striving to identify and eradicate undesirable, discriminative and possibly unlawful aspects of its operation. %
Here, we backtrack on this fixed model assumption to propose and explore a novel definition of \emph{cross-model fairness} where individuals can be harmed when one predictor is chosen ad hoc from a group of equally well performing models, i.e., in view of utility-based \emph{model multiplicity}. %
Since a person may be classified differently across models that are otherwise considered equivalent, this individual could argue for a predictor granting them the most favourable outcome, employing which may have adverse effects on other people. %
We introduce this scenario with a two-dimensional example and linear classification; then, we present a comprehensive empirical study based on real-life predictive models and data sets that are popular with the algorithmic fairness community; finally, we investigate analytical properties of cross-model fairness and its ramifications in a broader context. %
Our findings suggest that such unfairness can be readily found in real life and it may be difficult to mitigate by technical means alone as doing so is likely to degrade predictive performance.%
\end{abstract}


\maketitle

\begin{mygreenbox}
\footnotesize%
\textbf{Highlights}%
\begin{itemize}[topsep=0pt,label=\faLightbulbO,leftmargin=.5cm,itemindent=0cm,labelwidth=.5cm,labelsep=0cm,align=left]%
    \item Cross-model fairness guarantees that a person receives the same prediction for a collection of classifiers with identical or comparable predictive performance, i.e., under utility-based model multiplicity.%
    \item When at least one prediction output by multiple equivalent models for a single individual is perceived as favourable, the person might argue for the precedence of this outcome.%
    \item Cross-model fairness is consistent with the Blackstone's ratio and ``presumption of innocence'' -- lack of convincing evidence ought to warrant the most favourable treatment.%
    \item Granting each person the most favourable decision afforded by a collection of equivalent models may degrade the overall predictive performance on the underlying task, especially when the employed family of models is highly expressive.%
    \item A possible solution is to limit the number of admissible predictors by imposing appropriate modelling \emph{restrictions} that are consistent with the social context and the natural process governing the generation of the underlying data, as well as to employ a \emph{justified} prediction aggregation strategy across classifiers with identical or comparable performance.%
\end{itemize}
\end{mygreenbox}
\vspace{.285em}%
\begin{mybluebox}
\footnotesize%
\noindent\faGithub\hspace{.2cm}\textbf{Source Code}\quad%
\coderepo%
\end{mybluebox}
\vspace{.285em}%
\begin{myvioletbox}
\footnotesize%
\noindent\faFileTextO\hspace{.2cm}\textbf{Published in}\quad%
ACM Journal on Responsible Computing (\href{https://doi.org/10.1145/3677173}{10.1145/3677173})%
\end{myvioletbox}

\clearpage\newpage

\section{Cross-model Fairness: A New Notion of Algorithmic Fairness in Machine Learning\label{sec:intro}}%

Data-driven predictive models are making impressive strides across numerous domains, leading to their proliferation throughout businesses and society, where they either support decision-making or outright automate relevant tasks. %
This speedy adoption of artificial intelligence (AI) and machine learning (ML) algorithms, however, outpaces research investigating the potential harm of these techniques across different aspects of everyday life. %
While excluding human operators from the decision-making process bears the promise of faster, more precise as well as consistent and replicable outcomes that lack implicit human biases, pre-existing (historical) patterns concealed in training data may easily overshadow these benefits. %
The ubiquity of oftentimes erratically and narrowly tested and validated models can therefore contribute to and amplify problems with fairness, accountability and robustness of the predictive tasks being addressed. %
This is of particular concern when AI and ML models affect people -- through possibly long-term or legally binding decisions -- which has been documented in various contexts including banking, school admission, job screening and judiciary ruling~\cite{o2016weapons,propublica2016machine}. %
It is thus critical, among other desiderata, to ensure fairness of the resulting predictions with respect to protected characteristics, e.g., ethnicity or gender, when dealing with both individuals and population subgroups.%

While mitigating disparate treatment of groups and individuals in view of protected attributes is of paramount importance, other notions of algorithmic fairness should not be neglected. %
Here, we explore its novel conceptualisation -- which we call \emph{cross-model fairness} -- where instead of focusing on bias exhibited by a single model we deal with their collection characterised by equal (or comparable) predictive performance according to a given evaluation strategy and metric~\cite{mcallister2007model,marx2020predictive}. %
Our notion of fairness is intended to engender trust in automated decision-making by guaranteeing that each individual is offered the most favourable outcome they can get given a fixed family of models pre-selected to solve the underlying predictive task. %
A specific scenario where cross-model fairness plays an important role is credit scoring; %
it is not uncommon for such models to be refined and tuned, yielding different realisations of the chosen modelling pipeline, which may affect the consistency of individual decisions. %
From a technical perspective, such a conceptualisation of fairness can be viewed as a notion of predictive \emph{robustness} when the model in question evolves while being constrained by its overall level of predictive performance. %

As a simple example of cross-model fairness consider the classification scenario depicted in Figure~\ref{fig:linera_model_multiplicity} shown in Section~\ref{sec:background} where three distinct predictors from the family of linear classifiers achieve 100\% \emph{accuracy} with respect to the displayed \emph{validation set}. %
While models perfect in this respect do not make any \emph{observable} mistakes, they may still suffer from \emph{disputable regions} within which they disagree, giving rise to possible claims of cross-model unfair treatment voiced by previously unseen individuals residing in these spaces. %
Whenever we cannot guarantee perfect classification on the dedicated validation data, these considerations become more immediate as certain individuals from within this set may be treated differently across apparently equivalent models, as seen in Figures~\ref{fig:linera_model_multiplicity_1} and \ref{fig:linera_model_multiplicity_2}. %
Notably, as we move towards more \emph{expressive} families of predictive algorithms, their complexity and parameterisation space grow, possibly making the observed phenomenon more prominent, especially for workflows whose optimisation is \emph{greedy} or \emph{stochastic}.%

From the viewpoint of predictive performance there may be no objective reason to favour one model over another. %
This observation extends to \emph{confusion matrices} -- also known as \emph{contingency tables} -- since in certain scenarios they offer identical classification summaries despite distinct errors being made (as shown later in Table~\ref{tab:contingency}). %
In this setting a predictor could possibly be selected at random, however %
such an arbitrary choice %
poses ethical dilemmas as it %
may be dismissed by (adversely affected) individuals who are classified differently across this collection of models. %
Specifically, they can argue to be treated with the predictor granting them the most favourable outcome (especially if the selection procedure was ad hoc in the first place). %
Nonetheless, as the models under consideration are intrinsically different, many choices will inevitably result in some people benefiting at the expense of others, which in any case is difficult to justify. %
Even if a collection of models boasts perfect performance with respect to a designated validation set, we still have to account for any disputable regions that are not covered by these data points, hence remain unobserved, yet allow individuals who are placed therein to claim unfair treatment in view of model multiplicity. %
This line of reasoning becomes especially important in the context of criminal justice, where a person might be entitled to be processed by the most advantageous model as long as it achieves certain performance for the (validation) population as a whole. %
This concept of fairness can be linked to the Blackstone's ratio~\cite{blackstone1830commentaries} or the ``presumption of innocence''~\cite{assembly1948universal}, where lack of convincing evidence (or a unanimous vote) warrants the most favourable treatment: ``It is better that ten guilty persons escape than that one innocent suffers.''%

Such observations ought to encourage developers of predictive models to implement various selection heuristics and criteria that account for properties beyond predictive performance~\citep{black2022model}; e.g., %
overall complexity or coverage of a model may be considered to increase the likelihood of its uniqueness, and classification with a reject or abstain option can be adopted to mitigate inconsistent predictions across a collection of models~\cite{chow1957optimum,chow1970optimum}. %
Another concern are cases where only relatively few classifiers that exhibit a given level of performance present an individual with a favourable outcome. %
For example, consider such a curated ensemble of predictive models taking the role of jurors where an overwhelming majority offers the less desirable decision. %
While the unfavourable ratio can be ignored with just a single model creating a precedent for a positive prediction, accounting for the proportion of classifiers and individual reasons for their respective outputs may provide important insights. %
However, with an abundance of predictive pipelines composed of diverse data processing techniques and model families, deriving bounds on the least and most favourable treatment of each individual within a dedicated \emph{fairness} validation data set may be impractical or even infeasible. %
Understanding the stability of each prediction under certain modelling assumptions -- even if limited to an empirical analysis -- could nonetheless enhance, or even guarantee, fairness, trustworthiness, accountability and robustness of important data-driven decisions, or delegate them to a human supervisor if deemed necessary. %

Fundamentally, cross-model fairness aims to encourage model developers to account for non-uniqueness of data-driven predictive algorithms that are built and optimised primarily to maximise utility, especially that the scope of multiplicity has been show to expand with increase in predictive performance~\citep{black2022model}. %
Once this phenomenon is acknowledged, addressing it is still a challenge given the proliferation of black boxes whose internals and operations can neither be understood nor scrutinised, often even by their engineers, thus offering no rationale for preferring one over another and leading to an arbitrary choice. %
Embracing this \emph{crisis of justifiability}, rather than ignoring it, is a first step towards ensuring \emph{due process} as well as \emph{equal opportunity} and \emph{fair access to resources} -- by eradicating (unintentional) individual or group favouritism -- which in high stakes domains such as lending is a legal requirement~\citep{black2022model}. %
While our paper predominantly explores technical solutions and adopts the strictest possible definition of cross-model fairness -- where every individual is entitled to the best possible prediction offered by a collection of performance-equivalent models -- this setting is not necessarily the most suitable across the board. %
Depending on the use case, we can rely on additional operational desiderata (e.g., to align predictors with social values and ethical norms), high-level model selection heuristics (e.g., to promote transparent and robust predictors), and alternative prediction aggregation mechanisms (e.g., to relax the notion of cross-model fairness), all of which complement the technical means that target model underspecification. %
Employing these strategies and disclosing the rationale for their particular implementation can thus protect model creators from claims of cross-model unfairness and allow them to operationalise this concept in practice; for example, in cases where it is justified to select a model at random, use majority voting, or draw the decision according to the distribution of individual predictions. %

To the best of our knowledge such considerations found limited recognition in fairness literature~\citep{black2022model}, albeit they have been reported in machine learning~\cite{marx2020predictive,watson2023predictive,hsu2022rashomon}. %
Our contributions aim to establish solid foundations as well as provide initial empirical investigation and theoretical analysis of \emph{prediction-based individual fairness under model multiplicity}, which we call \emph{cross-model fairness}, thus fill the aforementioned research gap~\cite{breiman2001statistical}. %
Specifically, this paper discusses relevant literature and positions our novel conceptualisation of \emph{cross-model fairness} within its scope (Section~\ref{sec:related_work}). %
Next, it formally defines our notion of fairness based on \emph{disputable regions} that arise under two distinct types of \emph{utility-based model multiplicity} stemming from either a \emph{strict} or \emph{relaxed} predictive performance criterion (Section~\ref{sec:background}). %
These findings outline concepts fundamental to cross-model fairness. %
We introduce them by investigating diverse crisp binary classification scenarios where one outcome is universally preferred over the other, e.g., grant or decline parole, for a two-dimensional synthetic data set, nonetheless, as we show later, our contributions generalise beyond this restrictive setting. %
In the process, we uncover caveats and assumptions relevant to both the modelling task and the approach used to evaluate predictive performance; %
these findings help us to identify the core principles of cross-model fairness under utility-based predictive multiplicity and devise strategies to address the (unintended) consequences of this phenomenon. %

After establishing these foundations, the paper proceeds to a comprehensive, large-scale, empirical study of cross-model fairness on three real-life data sets popular in the literature: Credit Approval, German Credit and Adult (Section~\ref{sec:experiments}). %
Our experiments are based on top-performing classification workflows published in the OpenML repository~\cite{vanschoren2013openml} -- a setup that captures a realistic diversity of models that arise naturally in the machine learning pipeline instead of being artificially created with the sole purpose of studying ramifications of model multiplicity -- which ensures credibility of our results. %
We publish an open source implementation of our experiments on GitHub\footnote{\coderepo} to enable their reproducibility and allow others to easily apply our evaluation protocol to the wide selection of data sets available in OpenML or elsewhere. %
We report cross-model fairness with two model multiplicity metrics -- \emph{discrepancy} and \emph{ambiguity}~\cite{marx2020predictive} -- showing the prevalence and severity of this phenomenon. %
To support this analysis and %
help to uncover the degree of unfairness across the multiplicity spectrum -- both as a high-level overview and an in-detail, instance-specific perspective -- we design and present a novel visualisation toolkit. %
We also study the utility of the fair-by-design ensemble model achieved by allowing an individual to choose the (most favourable) classifier. %
Our results show the detrimental effects of such an approach %
on the overall predictive performance of the underlying classification system, particularly so when the employed model is relatively expressive, therefore its decision boundary is flexible. %

Next, we offer an analytical treatment of cross-model fairness (Section~\ref{sec:multiplicity}), investigating the influence of the expressiveness (flexibility) of predictive models on their fairness, and the consequences of granting each individual the best possible outcome using the fair-by-design ensemble model. %
Specifically, we observe that achieving this notion of fairness may require %
limiting the number of admissible predictors by imposing upon them (modelling) restrictions consistent with the natural process governing the generation of the underlying data, and devising high-level model selection heuristics. %
The former strategy is in line with strong notions of ante-hoc interpretability that are necessary for predictive models deployed in high stakes domains~\cite{rudin2019stop,sokol2023reasonable}. %
We also discuss weaker notions of cross-model fairness that rely on prediction aggregation approaches other than always outputting the most favourable outcome, e.g., majority voting, selecting a model at random or drawing the decision according to the distribution of the individual predictions given by performance-equivalent models. %
Finally, we conclude our work and outline future research directions (Section~\ref{sec:conclusion_future}). %

To the best of our knowledge, we are the first to empirically study the extent and severity of cross-model unfairness for realistic classification workflows in view of predictive multiplicity. %
Other pieces of research in this space offer theoretical or algorithmic contributions that operate in quite restrictive settings, which limits their real-world impact. %
In contrast, our work: %
\begin{enumerate}
    \item provides a flexible definition of fairness in view of model multiplicity, called cross-model fairness; %
    \item reports results of a large-scale, comprehensive, empirical study of this phenomenon for real-life data sets and predictive workflows; %
    \item demonstrates how na\"ive enforcement of cross-model fairness in the form of the fair-by-design ensemble model may adversely impact predictive performance; %
    \item analyses and discusses the ramifications of such a notion of fairness; and %
    \item provides an open source implementation of our evaluation pipeline and a collection of graphical investigative tools to allow others to easily replicate our results and apply the same analysis to different data sets and model collections. %
\end{enumerate}

\section{Related Work\label{sec:related_work}}

Fairness of artificial intelligence and machine learning algorithms has attracted considerable attention in recent years following the proliferation of data-driven automated decision-making systems across real-life applications~\cite{o2016weapons}. %
Two main themes dominate this research field: \emph{individual} and \emph{group} fairness, both focusing on strategies to identify and mitigate disparate impact that manifests itself through discriminatory behaviour and diverse biases~\cite{barocas2017fairness}. %
Nonetheless, complementary viewpoints also emerge, investigating this topic under substantially different assumptions such as social and population changes over time, which brings to light considerations of the delayed impact of enforcing fair decisions~\cite{liu2018delayed}. %
Regardless, a nearly universal assumption found in the literature is to presuppose a \emph{fixed predictive model}, thus overlooking its provenance, evolution as well as any implicit choices related to it. %
This is at odds with the model \emph{multiplicity} phenomenon~\cite{breiman2001statistical} -- i.e., possible existence of a \emph{collection} of equally capable classifiers -- which has been largely neglected by the fairness community~\cite{fisher2019all,marx2020predictive,black2022model}.%

\emph{Individual fairness} deals with disparate treatment of a single person based on selected comparison criteria such as relevant \emph{protected characteristics}. %
This phenomenon can be captured by a dedicated \emph{similarity metric} -- assuming an intuitive notion that similar people should be treated comparably -- however defining such a measurement strategy is non-trivial~\cite{dwork2012fairness}. %
An alternative approach to individual fairness is formulated via \emph{counterfactuals}, whereby changing a (protected) attribute should not yield a different prediction. %
This framing is also very intuitive and relatively easy to test, however the concept of a ``counterfactual individual'' may be ill-defined. %
For example, altering ethnicity while preserving the remaining features intact may not be representative given that all these other personal traits and attributes are (implicitly) influenced by the single personal characteristic being manipulated. %
Notably, causality may be employed to partially alleviate such shortcomings~\cite{kusner2017counterfactual}. %
\emph{Group fairness}, on the other hand, deals with disparate treatment of sub-populations determined, for example, by partitions drawn across protected attributes, in which case parity may be achieved by tweaking the underlying model separately for each group, e.g., via their respective classification thresholds~\cite{hardt2016equality}. %
Importantly, many such notions of fairness are inherently incompatible~\cite{kleinberg2017inherent}, and enforcing some of them may require trading off a degree of predictive performance~\cite{friedler2021possibility}.%

Multiplicity of data-driven predictive models is a well observed phenomenon -- sometimes called the Rash\=omon effect of statistics -- where a group of classifiers exhibits comparable utility despite intrinsic differences~\cite{breiman2001statistical}. %
Among others, this non-uniqueness of models may result from distinct: %
\begin{enumerate*}[label={(\roman*)}]%
    \item families of employed predictors, e.g., linear models, decision trees and neural networks, %
    \item subsets of used attributes, %
    \item compositions of training data sets, %
    \item patterns found in data, or %
    \item model parameters, e.g., stemming from random initialisation, %
\end{enumerate*} %
with some arguing that exogenous information may be necessary to narrow down the scope of predictors with equivalent performance~\cite{mcallister2007model,fisher2019all,black2022model}. %
\emph{Procedural} multiplicity is the broadest category of this phenomenon; %
it encompasses all models with comparable utility whose internals differ, leading them to produce disparate decision surfaces or arrive at (possibly identical) predictions in unique ways~\citep{black2022model}. %
\emph{Predictive} multiplicity is a special case of procedural multiplicity where the models in question yield different predictions for some instances; it is usually evaluated on a dedicated test set. %
More generally, multiplicity can be linked to the concept of \emph{epistemic uncertainty} in machine learning, also known as \emph{systemic uncertainty}, which refers to lack of knowledge about the most optimal model~\cite{hullermeier2021aleatoric}. %
This situation leads to multiple admissible predictors, choosing among which is non-trivial; however, this type of uncertainty can be \emph{reduced}, e.g., by collecting more data or introducing additional modelling assumptions. %

The ramifications of multiplicity have only recently been studied for fairness, accountability and transparency of machine learning models. %
\citet{pawelczyk2020counterfactual} explored the implications of this phenomenon on veracity of counterfactual explanations, which property may be lost when switching between different models whose performance is comparable. %
A recent empirical study of multiplicity in ML explainability has shown the prevalent inconsistency of attribution methods across a range of different modelling scenarios~\cite{muller2023empirical}. %
In relation to fairness, model multiplicity has been used to understand the dependence of (inaccessible) classifiers on selected (protected) attributes, which can help to robustly identify discriminatory practices~\cite{fisher2019all}. %
It was also suggested as an additional metric for evaluating accountability of classifiers; %
specifically, \citet{marx2020predictive} showed how multiplicity may be problematic (from ethical and fairness standpoints) when individuals receive conflicting predictions, and offered to address this issue by granting them the most favourable decision. %
\citet{black2022model} presented a similar perspective, pointing out that multiplicity can mask \emph{intentional} discrimination -- by choosing a performance-equivalent model that assigns a desired, unfavourable output to selected individuals -- which unfairness can either be perpetuated on an \emph{individual} (local) or \emph{aggregate} (global) level, i.e., void individual or group fairness constraints across models from the multiplicity spectrum. %

While the line of reasoning outlined by \citet{marx2020predictive} is close to ours, their study is limited to linear models based on Mixed-Integer Programming -- %
restricting the modelling setting allows the authors to derive theoretical results and propose algorithmic solutions, nonetheless it curtails real-life applicability of their findings. %
Our work, in contrast, is rooted directly in fairness, thus provides a more fundamental and general treatment of model multiplicity in this context. %
We complement these contributions with a comprehensive analysis of real-life predictive workflows built for data sets commonly used in fairness studies, which constitutes the core of this paper. %
Our approach is unique when compared to related work in both the fairness and multiplicity domains in so far as it studies these topics for predictive models that arise naturally (in the OpenML environment) in contrast to \emph{artificial multiplicity} constructed specifically for research and experimentation purposes. %
We also use these empirical results to complement the theoretical and technical study of multiplicity as well as an analysis of its legal implications and a discussion of policy recommendations in view of individual and group fairness reported by \citet{black2022model}, which broadens the scope of our investigation. %

Fundamentally, cross-model unfairness may be abated (to an extent) by employing a predictive paradigm that is more complex than crisp (binary) classification. %
The previous section briefly discussed strategies that can be used to increase the chances of a model being unique, e.g., by enforcing various heuristics and criteria with respect to its properties other than predictive performance. %
Adopting \emph{classification with a reject or abstain option} is another approach to mitigate predictions that are unfair from a perspective of an individual in view of model multiplicity~\cite{chow1957optimum,chow1970optimum}. %
Dedicated techniques such as \emph{prediction intervals} or \emph{conformal predictions} can also be deployed. %
The former marks any instance falling within a specified area constructed around a decision boundary as unreliable, allowing it to be assigned the most favourable decision; %
the latter follows a similar strategy, instead outputting a set of admissible classes, which can be further processed as desired. %
A possible alternative is to use \emph{ensemble learning}, e.g., a random forest~\cite{breiman2001random}, where instead of a majority vote the prediction is derived through maximising the output of the contributing models. %
Machine learning algorithms with prediction margins, such as support vector machines~\cite{cortes1995support}, as well as probabilistic classifiers that offer confidence scores or trustworthiness measures are another option. %
While in most of such cases the acute signs of cross-model unfairness can be eliminated, the underlying problem may not necessarily be fully resolved, causing troubles down the line. %
This has been shown for probabilistic classification, where model multiplicity remains problematic~\cite{hsu2022rashomon,watson2023predictive}. %

\section{Navigating Model Multiplicity in View of Cross-model Fairness\label{sec:background}}%

The notion of fairness studied in this paper is built upon the \emph{model multiplicity} phenomenon understood here as existence of a collection of data-driven models that are indistinguishable in terms of their predictive performance under a fixed evaluation strategy~\cite{breiman2001statistical,marx2020predictive}. %
Furthermore, in this work we are interested in \emph{crisp binary classification} in which one outcome is universally preferred to the other by individuals whose case is being decided. %
Therefore, a model \(f : \mathcal{X} \mapsto \mathcal{Y}\) classifies an instance \(x \in \mathcal{X}\) as \(f(x) = \hat{y}\), where \(\hat{y} \in \mathcal{Y} \equiv \{0, 1\}\) and \(1\) represents a favourable outcome. %
Predictive performance of such a model is measured on a predetermined validation data subset \(\widetilde{X} = \{ x_i \}_{i=1}^n \) of size \(n \in \mathbb{N}\), where \( x_i \in \mathcal{X}\), accompanied by %
annotations \(\widetilde{Y} = \{y_i\}_{i=1}^n\), where \(y_i \in \mathcal{Y}\), using a selected metric \( m : \mathcal{Y} \times \mathcal{Y} \mapsto \mathbb{R} \) calculated as \( m \left( f(\widetilde{X}), \widetilde{Y} \right) \). %
In our initial analysis we rely on linear, polynomial, \(k\)-nearest neighbours and decision tree classifiers applied to a synthetic two-dimensional data %
set, which helps us to demonstrate the %
principles of cross-model fairness that arises under utility-based predictive multiplicity through visual inspection and hand-crafted examples; all of our results, nonetheless, can be easily generalised beyond this restrictive setting.%

\begin{figure}[b]
    \centering
    \begin{subfigure}[t]{.32\textwidth}
        \centering
        \includegraphics[width=\linewidth]{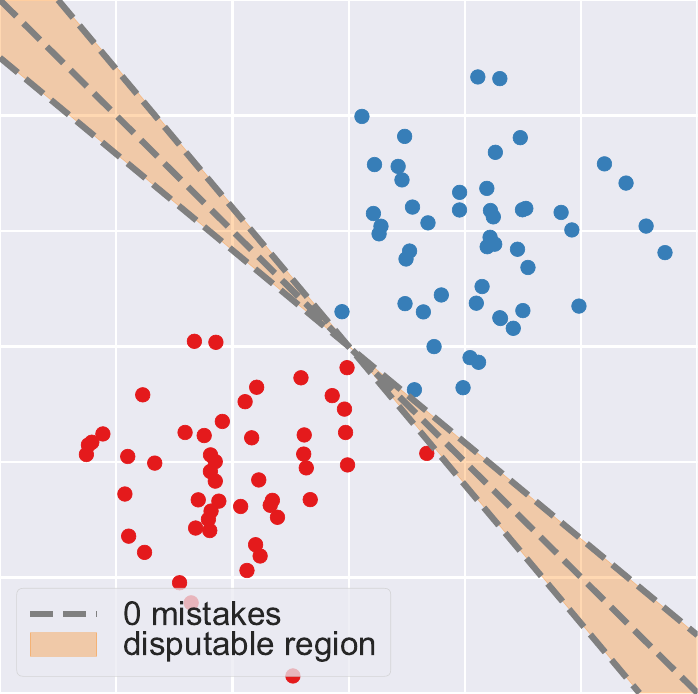}%
        \caption{No errors -- 100\% accuracy.\label{fig:linera_model_multiplicity}}%
    \end{subfigure}
    \hfill
    \begin{subfigure}[t]{.32\textwidth}
        \centering
        \includegraphics[width=\linewidth]{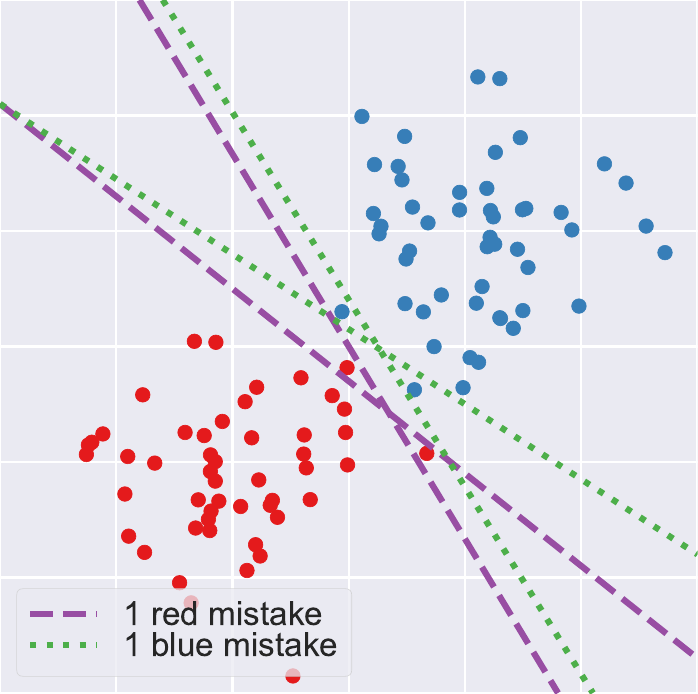}%
        \caption{One mistake -- 99\% accuracy.\label{fig:linera_model_multiplicity_1}}
    \end{subfigure}
    \hfill
    \begin{subfigure}[t]{.32\textwidth}
        \centering
        \includegraphics[width=\linewidth]{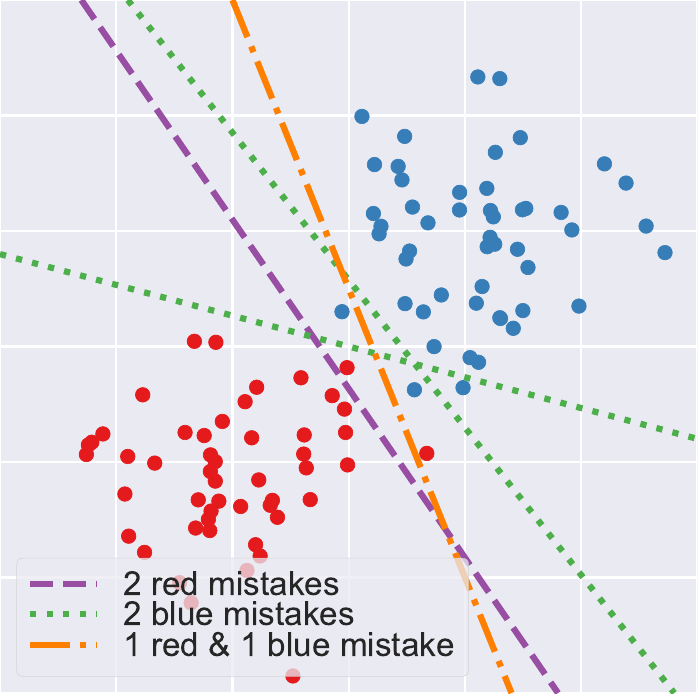}%
        \caption{Two errors -- 98\% accuracy.\label{fig:linera_model_multiplicity_2}}%
    \end{subfigure}
    \caption{Multiplicity of linear models with predictive performance (accuracy) measured on dedicated validation data (scatter plot). Perfect classifiers -- Panel~(\subref{fig:linera_model_multiplicity}) -- may still yield unfair decisions in view of model multiplicity for (initially unobserved) instances residing in disputable spaces. %
    Non-perfect models can make the same type of a mistake on different data points -- Panel~(\subref{fig:linera_model_multiplicity_1}) -- which may not be reflected in the chosen performance metric; %
    different types of a mistake -- Panel~(\subref{fig:linera_model_multiplicity_2}) -- may also go unnoticed in such a scenario.%
    \label{fig:linera_model_multiplicity_all}}%
\end{figure}

\begin{figure}[t]
    \centering
    \includegraphics[width=.32\textwidth]{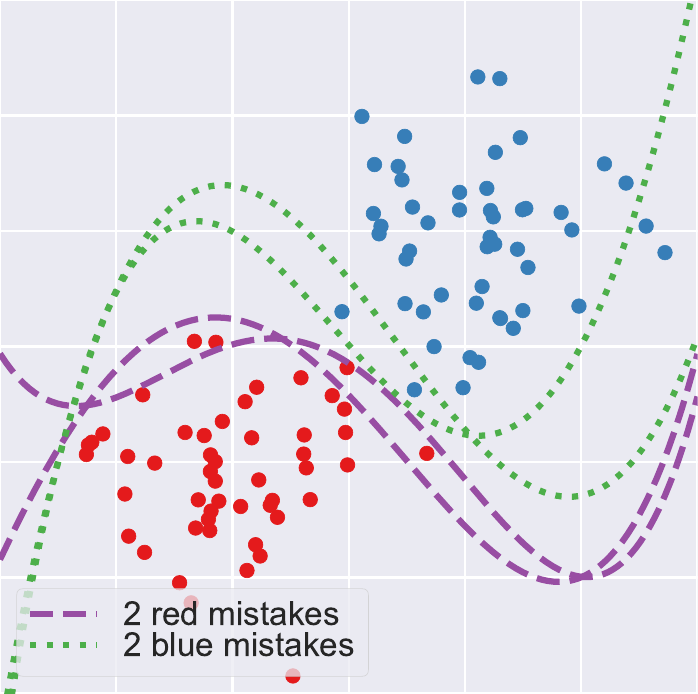}%
    \caption{Polynomial model multiplicity with two mistakes.\label{fig:polynomial_model_multiplicity_2}}
\end{figure}

While in principle model multiplicity may span a diverse range of classification models, we restrict our considerations to a confined \emph{family of predictive functions} -- also called a \emph{hypothesis class} -- inspired by such a formalisation of curves. %
This is desirable as distinct data-driven algorithms exhibit unique characteristics that translate into different shapes of their respective decision boundaries, allowing for diverse mistakes to be made, as seen when comparing Figures~\ref{fig:linera_model_multiplicity_2} and~\ref{fig:polynomial_model_multiplicity_2}. %
Therefore, a \textbf{family of data-driven models} \(\mathcal{F}\) consists of classifiers \(f \in \mathcal{F}\) based upon a single predictive algorithm and trained on a fixed data set, i.e., \(\mathcal{F} \subseteq \left\{f \; | \; f : \mathcal{X} \mapsto \mathcal{Y} \right\} \). %
It can further be constrained by imposing restrictions on the parameterisation space or optimisation approach, among other ML components, used with the selected method, e.g., a collection of decision trees no deeper than seven. %
Notably, algorithms whose training procedure is stochastic or greedy may yield distinct models from a single specification when run multiple times. %
A family of models can also, by extension, include more complex predictive workflows built with pre- or post-processing steps such as input normalisation or output calibration. %
It is therefore undesirable to further constrain or formalise our notion of data-driven model family given the diversity of predictive pipelines that can be derived from fundamental algorithmic building blocks. %

To address cross-model unfairness under predictive multiplicity we further identify a subset of classifiers from a single family that all share a fixed level of predictive performance (i.e., \emph{utility}). %
To this end, we need to select an evaluation metric; for simplicity, we employ accuracy throughout this paper, which is defined as \(m_{\text{acc}}(\hat{Y}, Y) = \sum_{\hat{y}, y \in \hat{Y}, Y} \mathds{1}_{\hat{y} = y} \; / \; |Y|\), where \(\hat{Y} = f(X)\) are predictions and \(Y\) are annotations for a data set \(X\). %
Moreover, we require a designated collection of labelled instances \((\widetilde{X}, \widetilde{Y})\) to serve as a dedicated \emph{predictive performance validation set}. %
This adds to the existing training and validation data used to fit the model and tune its hyperparameters, both of which constitute an integral part of any AI or ML workflow. %
Notably, the performance validation set \(\widetilde{X}\) can also be employed to evaluate our notion of fairness, however in certain cases -- reviewed in Section~\ref{sec:multiplicity} -- separating the two may be beneficial.%

\begin{figure}[t]
    \includegraphics[width=.23\textwidth]{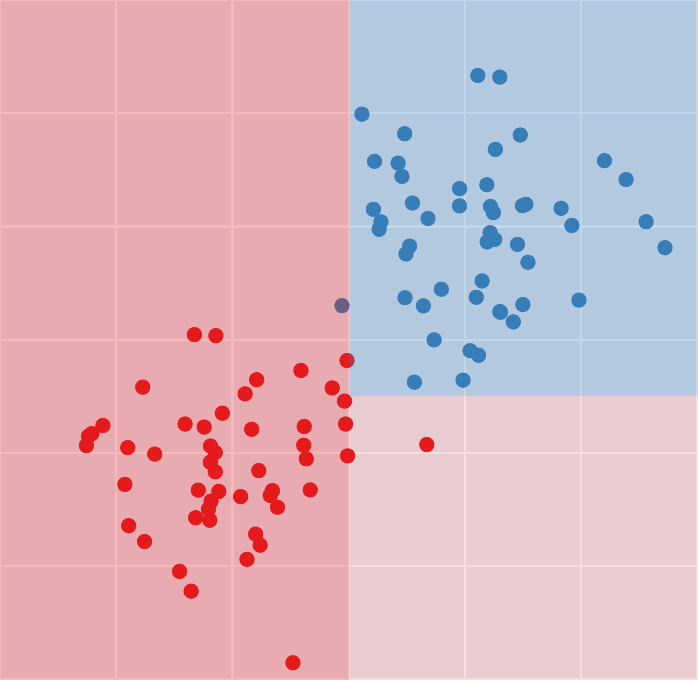}%
    \hspace{1em}%
    \includegraphics[width=.23\textwidth]{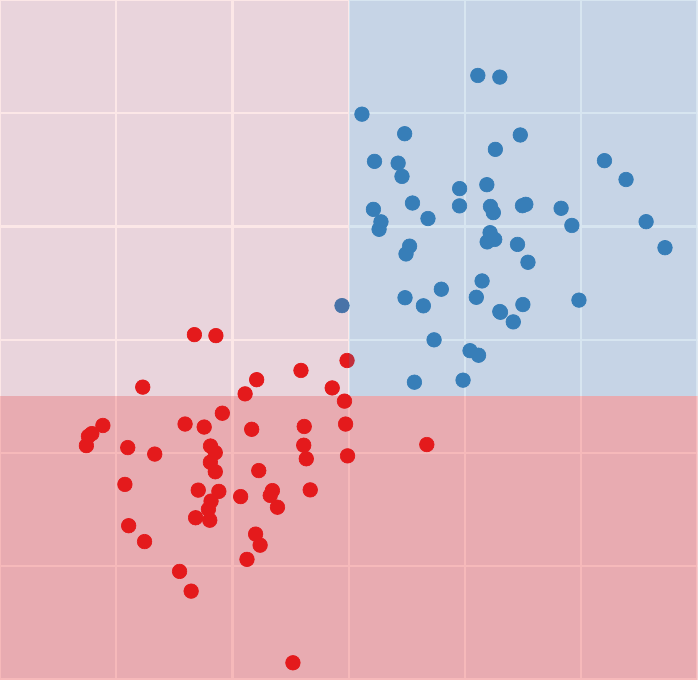}%
    \caption{Theoretical multiplicity of decision trees. While the models are distinct their predictions are identical for the entire data space.\label{fig:decision_tree_multiplicity}}%
\end{figure}

In this setting, %
\emph{utility-based model multiplicity} \(\mathcal{F}_\epsilon\) is determined by a fixed level of predictive performance \(\epsilon \in \mathbb{R}\) shared by classifiers from a single family of data-driven models \(\mathcal{F}\). %
In particular, there are two \emph{meaningful} viewpoints on multiplicity (which are compatible with its \emph{procedural} and expand its \emph{predictive} definitions given in Section~\ref{sec:related_work}):%
\begin{description}%
    \item [population-based] where the performance of each model is determined based on the entire data space \(\mathcal{X}\), which may be infeasible given a possibly infinite number of qualifying models (e.g., slight alterations of a linear classifier); and%
    \item [validation-based] where the performance is measured on a dedicated validation set \(\widetilde{X}\), making the problem tractable.%
\end{description}
In view of the former, the three models shown in Figure~\ref{fig:linera_model_multiplicity} are distinct, whereas based on the latter they are indistinguishable, i.e., they belong to a single multiplicity class \(\mathcal{F}_\epsilon\). %
Notably, either of these two notions is distinct from a purely \emph{theoretical non-uniqueness} of models within a single family, where the same decision boundary -- thus unobservable changes -- can be achieved with different realisations of a predictive pipeline; for example, see Figure~\ref{fig:decision_tree_multiplicity}, which depicts two structurally different decision trees that yield identical classification of the entire data space. %
Throughout this paper we are predominantly concerned with the validation-based multiplicity, which is outlined in Definition~\ref{def:multiplicity}. %
Such a setting simplifies our considerations since, in practice, it limits the number of models that we need to account for. %

\begin{definition}\label{def:multiplicity}
    \textbf{Utility-based model multiplicity} \(\mathcal{F}_\epsilon\) is captured by a set of classifiers from a single family of models \(\mathcal{F}\), all of which offer fixed predictive performance \(\epsilon\) for a given metric \(m\) %
    computed on a dedicated validation data set and its labels \((\widetilde{X}, \widetilde{Y})\):%
    \[
        \mathcal{F}_\epsilon \coloneqq \left\{ f \in \mathcal{F} : m\left( f(\widetilde{X}), \widetilde{Y} \right) = \epsilon \right\} \text{.}%
    \]
\end{definition}

\begin{table}[b]
    \centering
    \caption{Even though confusion matrices are different for the violet, orange and green classifiers shown in Figure~\ref{fig:linera_model_multiplicity_2}, they cannot convey the two distinct errors made by the green models on individual instances -- see Panel~(\subref{tab:contingency:green}) below.\label{tab:contingency}}%
    \begin{subtable}{0.30\linewidth}
    \centering
    \caption{Violet model.\label{tab:contingency:violet}}
    \begin{tabular}{rrr}
       \toprule
        & red & blue\\
       \midrule%
       red  & 48 & 0\\
       blue & 2  & 50 \\
       \bottomrule
    \end{tabular}
    \end{subtable}
    \hspace{0.03\linewidth}%
    \begin{subtable}{0.30\linewidth}
    \centering
    \caption{Orange model.\label{tab:contingency:orange}}
    \begin{tabular}{rrr}
       \toprule
        & red & blue\\
       \midrule%
       red  & 49 & 1\\
       blue & 1  & 49 \\
       \bottomrule
    \end{tabular}
    \end{subtable}
    \hspace{0.03\linewidth}%
    \begin{subtable}{0.30\linewidth}
    \centering
    \caption{Green models.\label{tab:contingency:green}}
    \begin{tabular}{rrr}
       \toprule
        & red & blue\\
       \midrule%
       red  & 50 & 2\\
       blue & 0  & 48 \\
       \bottomrule
    \end{tabular}
    \end{subtable}
\end{table}

It is important to note that the choice of the performance metric determines which models are considered equivalent. %
For example, all of the classifiers shown in Figure~\ref{fig:linera_model_multiplicity_2} make two mistakes, hence are indistinguishable with respect to \emph{accuracy}; however, a different metric, e.g., \emph{precision}, \emph{recall} or \emph{specificity}, would make the violet, orange and green models distinct. %
Since all of the predictive performance metrics for classification tasks are derived from confusion tables, instead of relying on calculated numerical values we can refer directly to the underlying contingency matrices. %
Table~\ref{tab:contingency} lists specific errors made for the red and blue classes by the models shown in Figure~\ref{fig:linera_model_multiplicity_2}, where the green predictors incorrectly classify two blue points, the violet model errs on two red examples and the orange classifier mistakes one blue and one red instance. %
However, even such a fine-grained approach is insufficient to capture the \emph{same type of a mistake made on different data points}, i.e., individual predictions, motivating our notion of cross-model fairness under utility-based multiplicity. %
We are therefore interested in classifiers that are indistinguishable in terms of performance measured on a dedicated validation data set, but provide distinct class assignment for certain instances. %
Our approach does not account for ground truth since the individuals affected by cross-model unfairness may not necessarily be included in the collection of instances used to assess this property, thus they may lack (reliable) labels. %

In its simplest form, cross-model fairness can be studied by defining a designated (unlabelled) \textbf{fairness validation data set} \(\mathring{X} = \{ x_i \}_{i=1}^m\) of size \(m \in \mathbb{N}\), where \( x_i \in \mathcal{X}\), through which we can identify inconsistent predictions across a selected collection of models. %
A generalisation of such individual mistakes -- inspired by the \emph{population-based} model multiplicity viewpoint -- are \emph{disputable spaces} \(\mathring{\mathcal{X}}_{\mathcal{F}_\epsilon}\), an example of which is shown in Figure~\ref{fig:linera_model_multiplicity}. %
This extension -- outlined by Definition~\ref{def:disputable} -- is useful when the selected fairness validation data set is sparse and cannot capture inconsistent predictions at the desired level of detail, i.e., \(\mathring{X} \cap \mathring{\mathcal{X}}_{\mathcal{F}_\epsilon} = \emptyset \), hence no instance from \(\mathring{X}\) is placed within the disputable spaces \(\mathring{\mathcal{X}}_{\mathcal{F}_\epsilon}\). %
Notably, the shape of these regions is influenced by the model family \(\mathcal{F}\); for example, consider the polynomial classifiers shown in Figure~\ref{fig:polynomial_model_multiplicity_2} for which such spaces are much more complex than in the case of linear classification (refer to Figure~\ref{fig:linera_model_multiplicity_all}). %
Logical models, on the other hand, constrain disputable regions to (hyper-)rectangles since they impose axis-parallel splits on the data space (see Figure~\ref{fig:decision_tree_multiplicity}).%

\begin{definition}\label{def:disputable}
    A \textbf{disputable space} (or region) \(\mathring{\mathcal{X}}_{\mathcal{F}_\epsilon} \subseteq \mathcal{X}\) for utility-based model multiplicity \(\mathcal{F}_\epsilon\) is given by%
    \[%
    \mathring{\mathcal{X}}_{\mathcal{F}_\epsilon} \coloneqq \left\{ x \in \mathcal{X} : \exists \; f_i, f_j \in \mathcal{F}_\epsilon \;\; \text{s.t.} \;\; f_i(x) \neq f_j(x) \right\}%
    \]
    for a chosen predictive metric \(m\) and labelled (performance) validation set \((\widetilde{X}, \widetilde{Y})\), where \(\forall \; f \in \mathcal{F}_\epsilon \;\; m\left( f(\widetilde{X}), \widetilde{Y} \right) = \epsilon\). %
    It is a generalisation of instance-based cross-model unfairness determined by a designated fairness validation data set \(\mathring{X}\), which is necessarily limited in scope. %
\end{definition}

If the main property considered while choosing a classifier \(f \in \mathcal{F}_\epsilon\) is predictive performance, then from the perspective of utility-based model multiplicity \(\mathcal{F}_\epsilon\) all such predictors may be viewed as equivalent, without any particular preference for a given classifier. %
Lacking some further, well-defined selection criteria, an arbitrary choice can nonetheless lead to unfair cross-model treatment of individuals given the existence of an equally suitable predictor that may provide these people with a more favourable outcome. %
For example, consider the green classifiers shown in Figure~\ref{fig:linera_model_multiplicity_2}; both of them have identical confusion matrices (Table~\ref{tab:contingency:green}) yet only one of the two borderline blue individuals may be assigned the preferred outcome -- the red class -- depending on the model choice. %
Framing such a scenario as an automated decision between granting (red) or denying (blue) parole in an (algorithmically-supported) judicial hearing, performance-based indistinguishability of models becomes an important (ethical) factor. %
While ideally the task should be to minimise the scope of any disputable region, i.e., to deal with \(\mathring{\mathcal{X}}_{\mathcal{F}_\epsilon}\), in this paper we focus on a designated fairness validation set \(\mathring{X}\), a setting formalised by Definition~\ref{def:fairness_definition}, which specifies cross-model fairness through predictive consistency of a model family for a fairness validation data set. %
Notably, such a notion of fairness can be expanded from individuals to \emph{groups} by comparing the impact of multiplicity, e.g., ratios of affected instances, across them (refer to \emph{individual} and \emph{aggregate} levels of unfairness discussed in Section~\ref{sec:related_work}). %

\begin{definition}\label{def:fairness_definition}
    A classifier \(f \in \mathcal{F}_\epsilon\) is \textbf{cross-model fair} towards individuals \(x \in \mathring{X}\) in view of utility-based model multiplicity \(\mathcal{F}_\epsilon\) iff %
    \(\; \forall \; f^\prime \in \mathcal{F}_\epsilon \; \forall \; x \in \mathring{X} \;\; f(x) = f^\prime(x)\).%
\end{definition}

\begin{figure}[t]
    \centering
    \includegraphics[width=0.32\textwidth]{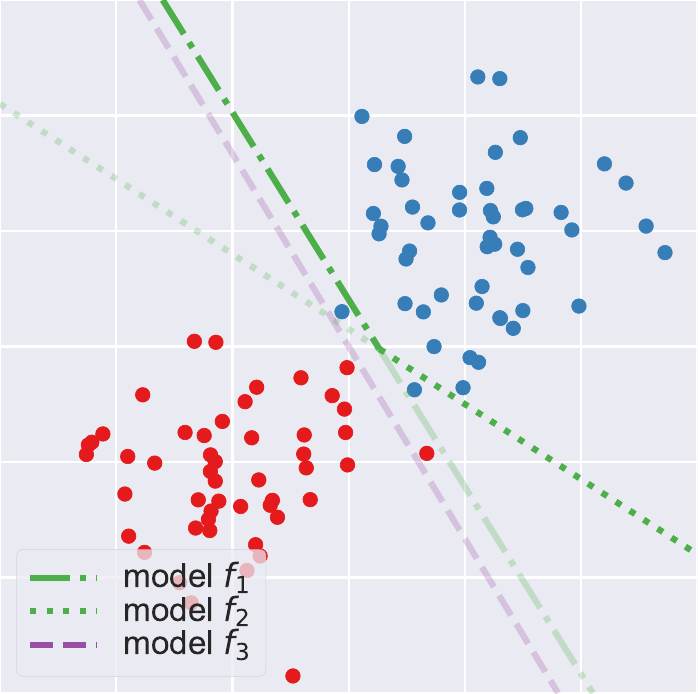}%
    \caption{
    Maximising the feature space predicted with the more favourable outcome (red) by %
    joining (at intersections) the two green and one purple (top left and out of view) models \(f \in \mathcal{F}_\epsilon\) shown in Figure~\ref{fig:linera_model_multiplicity_1} gives the cross-model individually fair classifier \(f^\star\) under utility-based predictive multiplicity \(\mathcal{F}_\epsilon\).%
\label{fig:linera_model_multiplicity_1_merged}}%
\end{figure}

In such a setting, eliminating unfairness caused by disparate decisions found across admissible models entails treating each individual with the most beneficial predictor \(f \in \mathcal{F}_\epsilon\). %
This observation leads us to define a new, fair-by-design model \(f^\star\) built by aggregating all of the classifiers that are equivalent under utility-based model multiplicity \(\mathcal{F}_\epsilon\). %
We do so by incorporating disputable regions into the decision space predicted with the more favourable outcome, therefore maximally growing its coverage. %
This approach may be considered as a simple model ensemble where we are interested in the best, rather than the average, result. %
Figure~\ref{fig:linera_model_multiplicity_1_merged}, for example, shows such a composition of classifiers for the models depicted in Figure~\ref{fig:linera_model_multiplicity_1}, assuming that the red class (\(1\)) is preferred to the blue class (\(0\)). %
The new decision boundary is constructed by joining linear models at their intersection -- the two green segments in the visible frame -- as to maximise the area predicted with the favourable outcome. %
Notably, it is possible that the fair-by-design model itself does not belong to the family, i.e., \(f^\star \notin \mathcal{F}\), which is captured by the aforementioned example -- the decision boundary of \(f^\star\) is not linear. %
The cross-model individually fair classifier \(f^\star\) is formalised in Definition~\ref{def:fair_model}. %
In theory, achieving fairness in such a way requires access to all of the classifiers that are equivalent under utility-based model multiplicity \(\mathcal{F}_\epsilon\) since we are concerned with disputable spaces. %
While this may be impractical or impossible, identifying cross-model unfairness for an individual is contingent on having access to at least one model whose prediction disagrees with another available classifier, allowing us to (iteratively) rectify this undesirable behaviour in practice. %
Alternatively, a representative sample of predictors -- either created intentionally for this purpose or available as a by-product of model development, optimisation and refinement -- could be used to this end. %

\begin{definition}\label{def:fair_model}
    The \textbf{cross-model individually fair classifier} \(f^\star\) under utility-based model multiplicity \(\mathcal{F}_\epsilon\) is defined as%
    \[
    f^\star(x) \coloneqq \argmax_{f \in \mathcal{F}_\epsilon} f(x)
    \]
    for any instance \(x \in \mathcal{X}\). %
    (Recall that \(1\) is the preferred prediction in our binary classification setting.)%
\end{definition}

Thus far we operated under strict utility regime \(\epsilon\), nonetheless this notion can be relaxed by allowing a certain deviation from the desired level of predictive performance. %
We can argue in favour of such an approach given that the utility is measured on a data subset, an expansion or contraction of which is likely to yield some variation in predictive performance. %
We can specify the allowed \emph{tolerance} through an additional parameter \(\delta \in \mathbb{R}^+\) -- where the performance band is denoted with \(\epsilon \pm \delta\) -- which extends the utility-based model multiplicity \(\mathcal{F}_\epsilon\) provided in Definition~\ref{def:multiplicity} to \(\mathcal{F}_{\epsilon\pm\delta}\) as outlined in Definition~\ref{def:multiplicity_delta}. %
An alternative operationalisation of this concept is to round the predictive performance \(\epsilon\) to a specified decimal point \(\delta\), written as \(\epsilon \simeq \delta\); for example, \(\epsilon\simeq10^{-2}\) indicates \(\epsilon\) rounded to the second decimal place. %
We use the latter approach throughout our experiments to streamline the exploration of real-life data sets.%

\begin{definition}\label{def:multiplicity_delta}
    \textbf{Relaxed utility-based model multiplicity} \(\mathcal{F}_{\epsilon\pm\delta}\) is a collection of classifiers from across a single family of models \(\mathcal{F}\) that exhibit a level of predictive performance within a fixed range \([\epsilon - \delta, \; \epsilon + \delta]\) for a chosen metric \(m\):%
    \[
        \mathcal{F}_{\epsilon\pm\delta} \coloneqq \left\{ f \in \mathcal{F} : \epsilon-\delta \leq m\left( f(\widetilde{X}), \widetilde{Y} \right) \leq \epsilon+\delta \right\} \text{,}%
    \]
    where (\(\widetilde{X}, \widetilde{Y})\) is a dedicated validation data set with labels, and \(\delta\) is the tolerance of the predictive performance level \(\epsilon\).%
\end{definition}

\section{Real-life Impact of Predictive Multiplicity on Cross-model Fairness\label{sec:experiments}}%

\begin{figure}%
    \centering
    \begin{subfigure}[t]{0.31\linewidth}%
        \centering
        \includegraphics[trim={0 -51.44pt 0 0},clip,width=.96\textwidth]{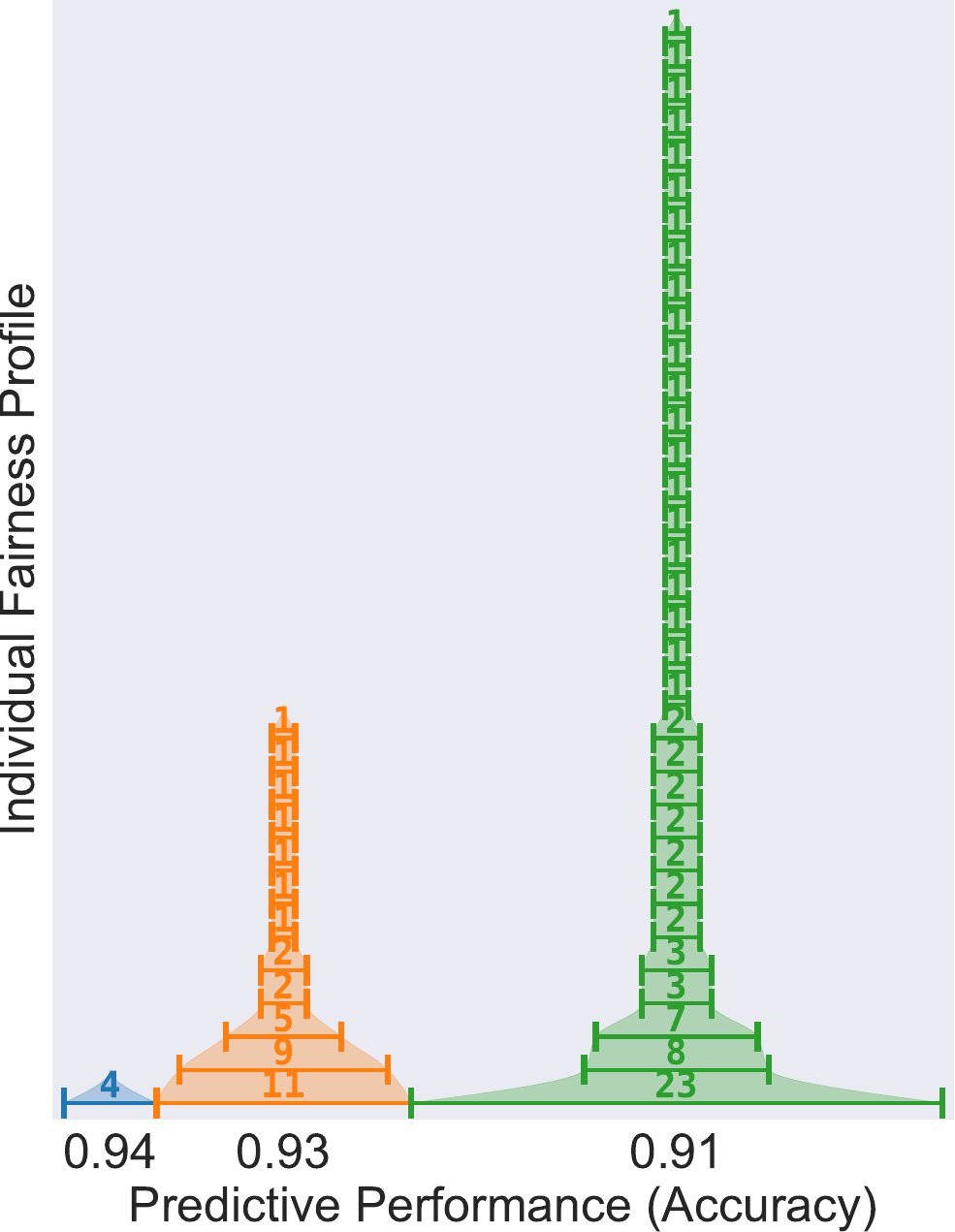}%
        \caption{Credit Approval -- top 3 performance bands spanning 119 models.\label{fig:experiments:stability_profile:credit_aproval}}%
    \end{subfigure}%
    \hfill%
    \begin{subfigure}[t]{0.31\linewidth}%
        \centering
        \includegraphics[trim={0 -51.44pt 0 0},clip,width=.96\textwidth]{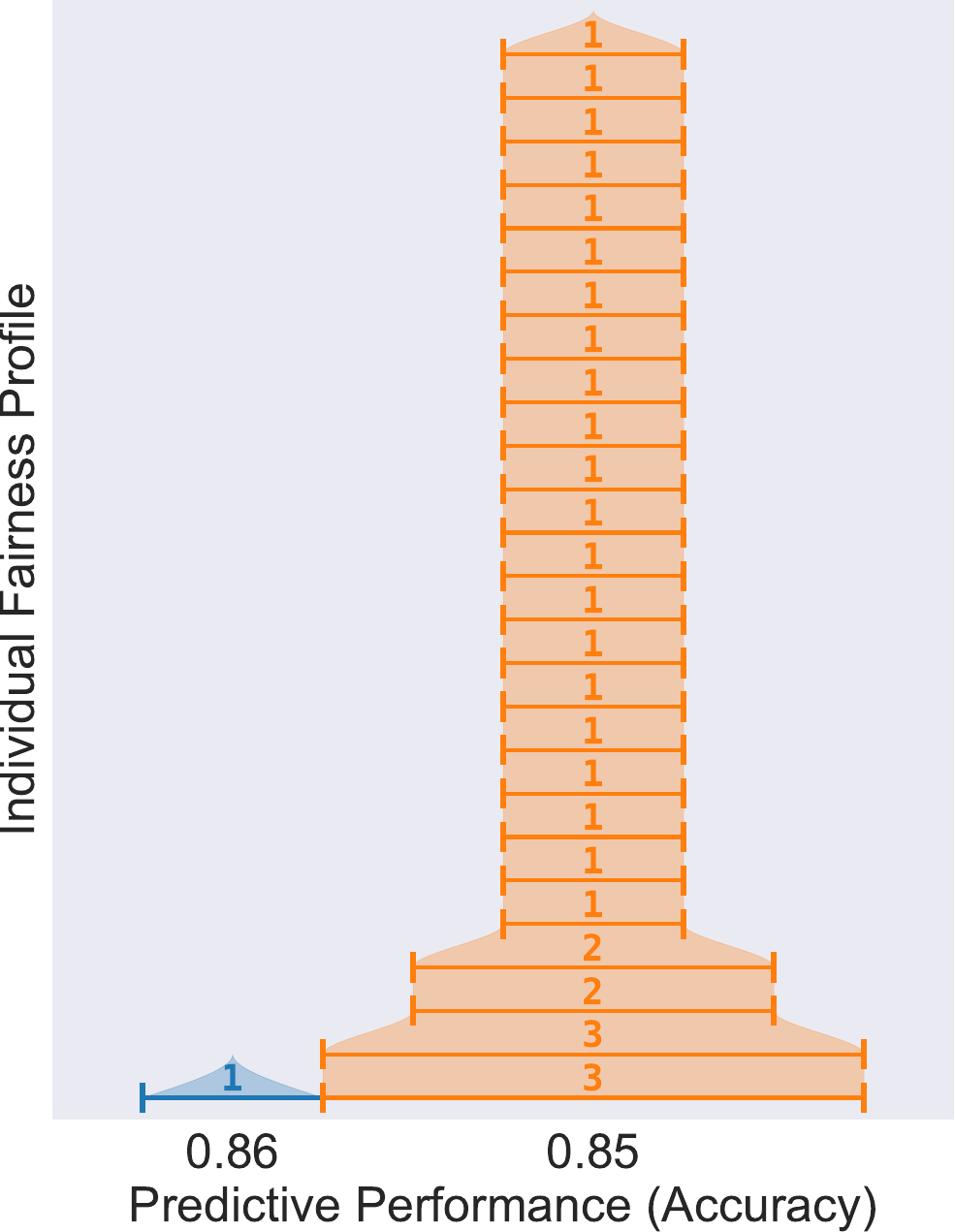}%
        \caption{German Credit -- top 2 performance bands spanning 32 models.\label{fig:experiments:stability_profile:german_credit}}%
    \end{subfigure}%
    \hfill%
    \begin{subfigure}[t]{0.31\linewidth}%
        \centering
        \includegraphics[width=.96\textwidth]{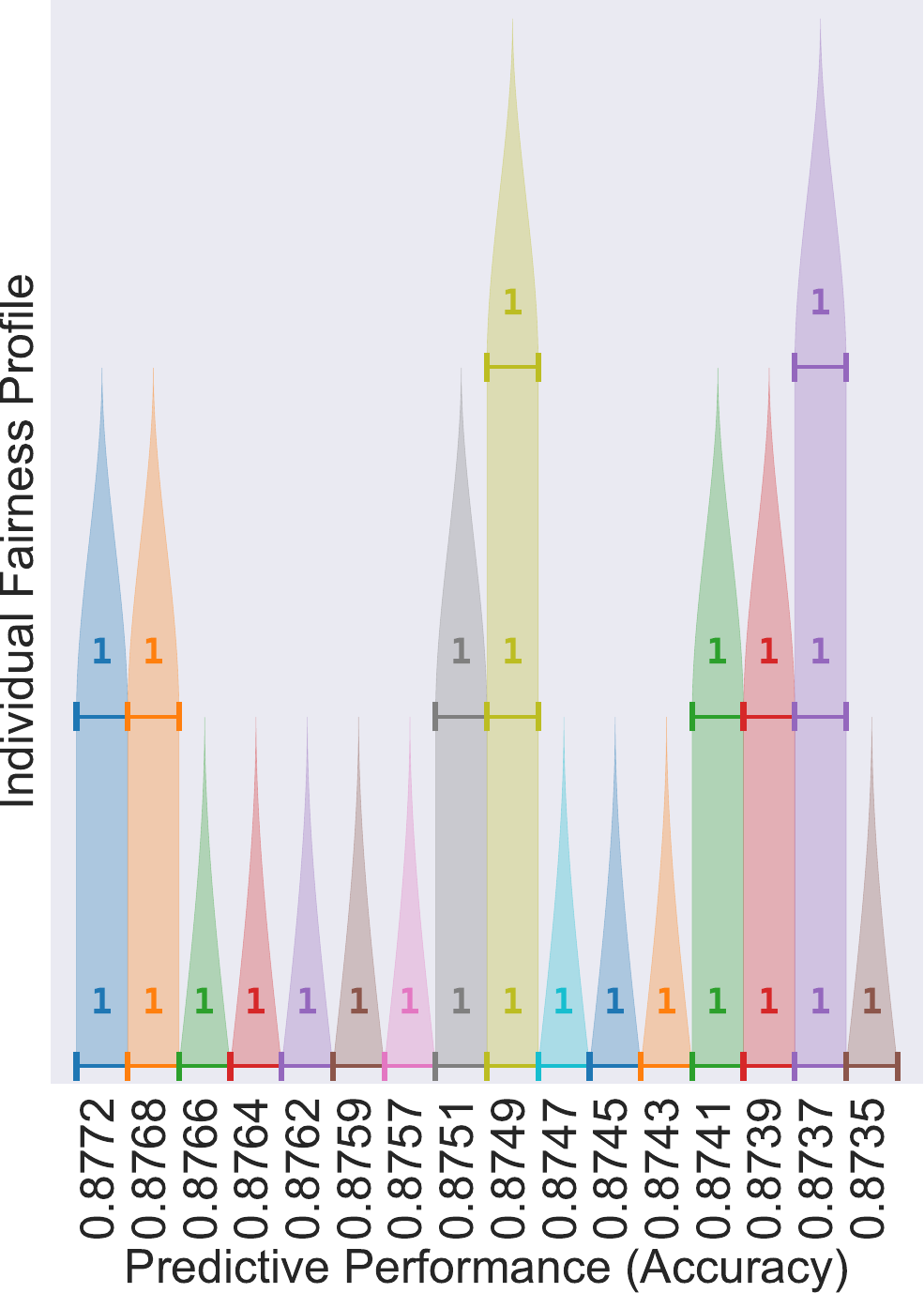}%
        \caption{Adult -- top 16 performance bands spanning 25 models.\label{fig:experiments:stability_profile:adult}}%
    \end{subfigure}%
    \caption{%
Stability profiles for the top \(n\) performance bands of each data set. %
This visual inspection tool displays the counts of unique prediction vectors, i.e., class assignments across the entire fairness validation set, for a collection of models from a chosen family \(\mathcal{F}\) grouped by predictive performance -- depicted as stacks of different colours -- measured on a dedicated validation set, in our case using accuracy. %
For example, the orange pyramid in Panel~(\subref{fig:experiments:stability_profile:credit_aproval}) reveals 36 models with 93\% accuracy distributed across 12 distinct prediction vectors -- the number of horizontal segments -- as shown by the reported counts and reflected in the width of each bar.%
\label{fig:experiments:stability_profile}}%
\end{figure}

\begin{figure}%
    \begin{subfigure}[t]{0.48\linewidth}%
        \centering
        \includegraphics[width=1.\textwidth]{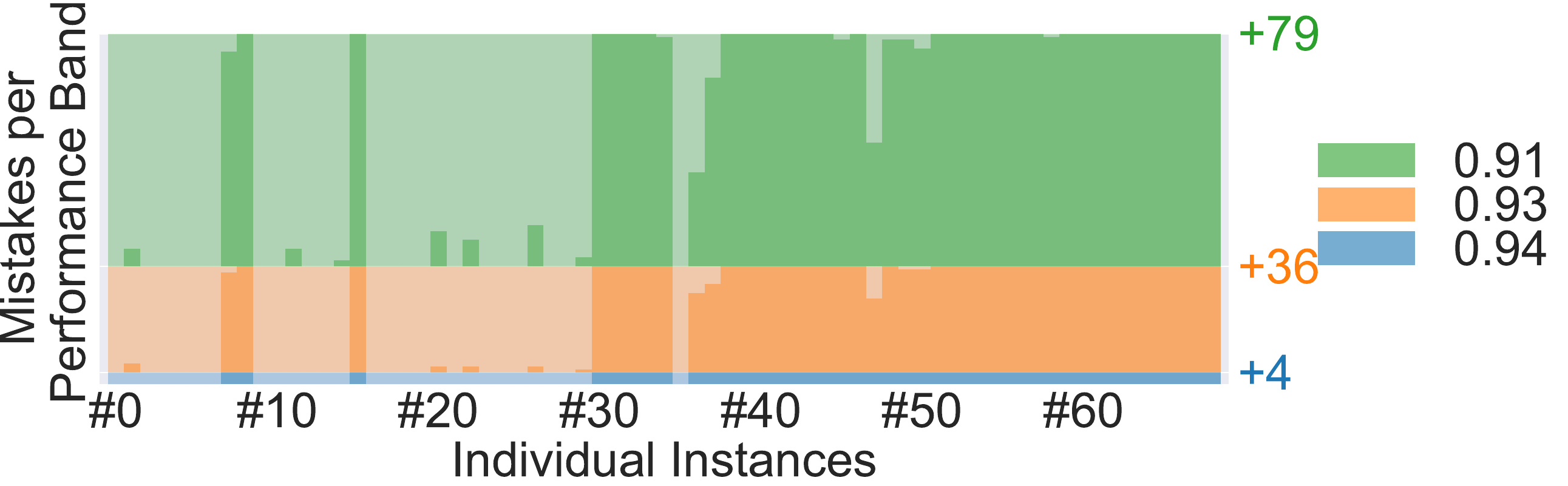}%
        \caption{Credit Approval -- top 3 performance bands with 69 instances.\label{fig:experiments:fairness_profile:credit_aproval}}
    \end{subfigure}
    \hfill
    \begin{subfigure}[t]{0.48\linewidth}%
        \centering
        \includegraphics[width=1.\textwidth]{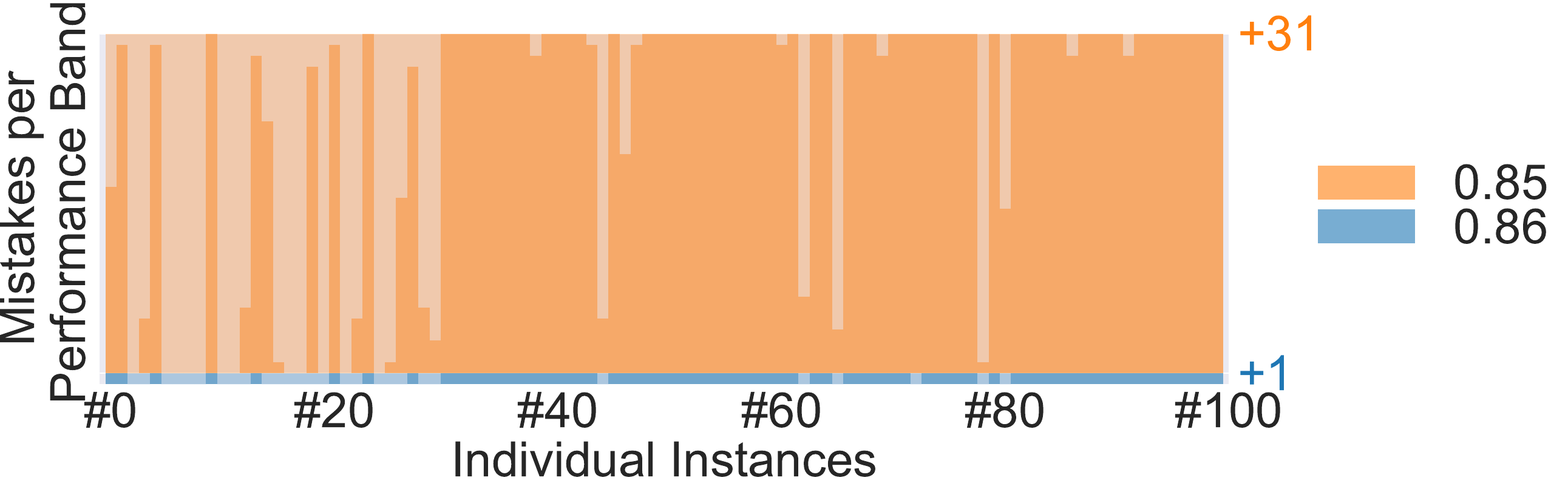}%
        \caption{German Credit -- top 2 performance bands with 100 instances.\label{fig:experiments:fairness_profile:german_credit}}
    \end{subfigure}
    \\[1em]%
    \begin{subfigure}[t]{0.96\linewidth}%
        \centering
        \includegraphics[width=.77\textwidth]{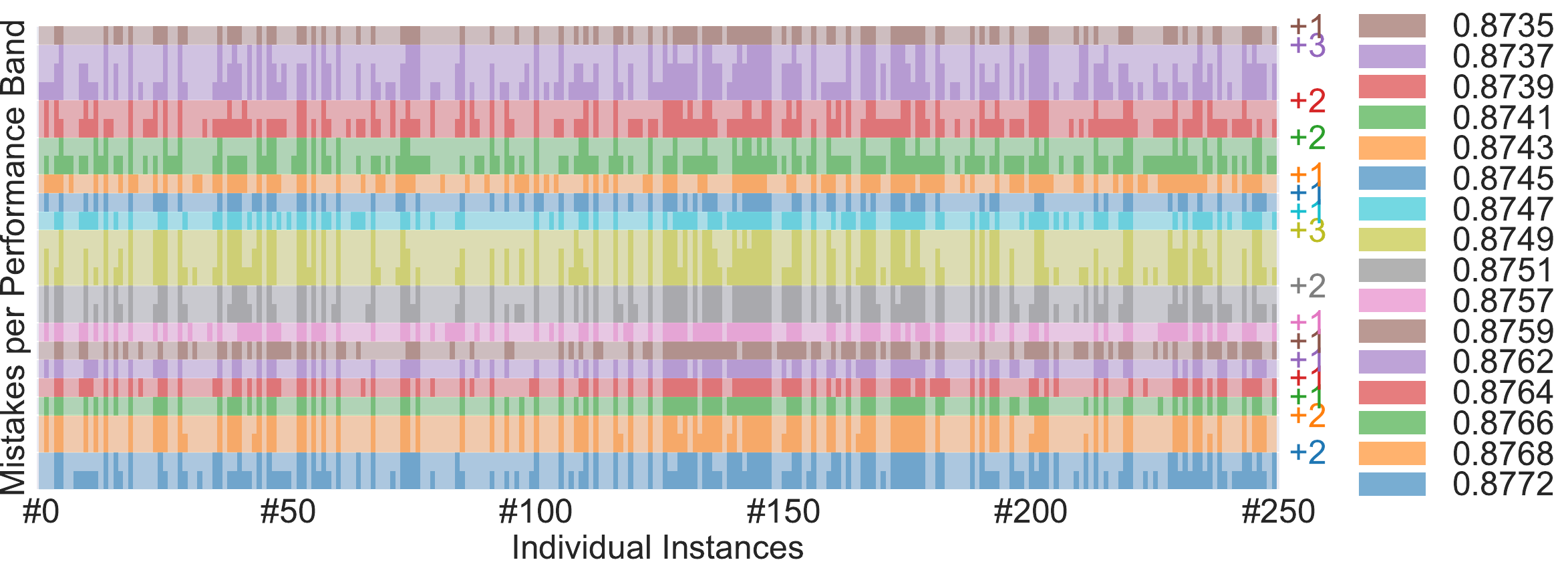}%
        \caption{Adult. For clarity, the profile shows a sample of 250 instances out of 586 individuals treated unfairly by any of the 25 models spanning the top 16 performance bands depicted in Figure~\ref{fig:experiments:stability_profile:adult}. There are a total of 4,885 data points in this validation set.\label{fig:experiments:fairness_profile:adult}}%
    \end{subfigure}
    \caption{%
Fairness profiles plotted for the chosen data sets. %
This visual inspection tool depicts the behaviour of individual predictive models -- spread across rows -- for each instance -- captured by a single column -- in the fairness validation set. %
The performance bands are colour-coded and the saturation of cells differentiates between the two possible predictions. %
The \emph{summary variant} of a fairness profile -- used here -- sorts the predictions vertically in every column, separately for each performance band, to improve readability of large fairness validation data sets, especially since they cannot be easily ordered. %
For example, Panel~(\subref{fig:experiments:fairness_profile:adult}) displays the behaviour of 25 models (rows) split across 16 (colour-coded) performance bands shown previously in Figure~\ref{fig:experiments:stability_profile:adult} for a selection of individual instances.%
\label{fig:experiments:fairness_profile}}
\end{figure}

Having explored the fundamentals of cross-model fairness under predictive multiplicity, we shift our attention to the real-life occurrence and impact of this phenomenon, which we investigate by studying pre-existing collections of ML pipelines. %
In particular, we focus on a common scenario when such considerations may arise, namely iterating over ML models during their development. %
To reconstruct this setting we turn to OpenML: a reproducibility repository for machine learning experiments~\cite{vanschoren2013openml}. %
Specifically, we concentrate on three real-life data sets popular in fairness research -- \emph{Credit Approval}, \emph{German Credit} and \emph{Adult} -- investigating the numerous executions of the top-performing binary classification workflow available therein for these data -- histogram-based gradient boosting classification tree, random classification forest and decision tree-based AdaBoost respectively. %
In each case, the ML task is (supervised) crisp binary classification run on 10-fold cross validation, thus every workflow yields 10 distinct training and test data sets, and the same number of models. %
When reporting \emph{stability} and \emph{fairness profiles} -- Figures~\ref{fig:experiments:stability_profile} and \ref{fig:experiments:fairness_profile} -- we select the model trained on folds 2--10 whose performance is evaluated on fold 1, which also serves as the fairness validation set. %
We provide these visualisations for only one fold since our investigation of \emph{ambiguity} and \emph{discrepancy} -- which are measures of multiplicity shown in Figure~\ref{fig:amb_dics} -- across \emph{all} the folds shows a consistent behaviour. %
A summary of this setup given as identification numbers (IDs) of OpenML data sets, tasks and flows is provided in Table~\ref{tab:openml}. %

\begin{table}[t]
    \centering
    \caption{OpenML workflow execution counts, i.e., the number of distinct models, as well as identification numbers (IDs) of data sets, tasks and flows used in our experiments.\label{tab:openml}}%
    \begin{tabular}{rrrrr}
        \toprule
        Data Set        & Data ID & Task ID & Flow ID & Execution Count\\
        \midrule%
        Credit Approval & 29      & 29      & 12736   & 442\\
        German Credit   & 31      & 31      & 6794    & 102,306\\
        Adult           & 1590    & 7592    & 6970    & 411\\
        \bottomrule
    \end{tabular}
\end{table}

Using OpenML in lieu of building, training and testing our own classifiers %
allows us to perform a first-of-a-kind analysis of multiplicity based on realistic predictive models that arise naturally instead of being artificially constructed with the sole purpose of studying this phenomenon. %
Such an approach makes our results more credible, especially that we choose the top-performing models available for each data set. %
To ensure a large sample of classifiers we opt for well-established data sets; %
while this means that we are unable to investigate more modern data sets that current fairness research relies on, such an analysis can be easily done in the future using the open source implementation of our evaluation framework. %
The code needed to reproduce our experiments and the results reported in this section is published on GitHub -- see Section~\ref{sec:intro} for the link. %

As we have observed earlier in Section~\ref{sec:background}, standard performance metrics derived from confusion matrices are insufficient to study the extent of individuals being classified inconsistently for a chosen family of models \(\mathcal{F}_\epsilon\) -- see Table~\ref{tab:contingency:green} for an example. %
To fill this gap we design a toolkit capable of: %
\begin{itemize}
    \item measuring the severity of this phenomenon, %
    \item identifying the affected instances, and %
    \item effectively communicating these findings at the desired level of detail. %
\end{itemize}
To this end, we employ the numerical measures of multiplicity -- \emph{ambiguity} and \emph{discrepancy} -- introduced by \citet{marx2020predictive}, and build bespoke graphical analytic instruments around them. %
Additionally, we develop a comprehensive visualisation tool that summarises the prediction structure for each individual performance band \(\epsilon\) of a model family \(\mathcal{F}\) -- a classifier perspective called \emph{stability profile}. %
We complement this approach with a fine-grained inspection plot that captures precise classification results for all data points across chosen performance bands \(\epsilon\) -- a prediction view named \emph{fairness profile}. %
We first introduce and explain these tools; next, we apply them to the chosen data sets and model families \(\mathcal{F}\) from the OpenML repository; finally, we interpret our findings and discuss their implications. %
We complement our analysis by %
    investigating the same selection of properties when the performance bands are relaxed (Definition~\ref{def:multiplicity_delta}), and %
    inspecting the change in utility (accuracy in our case) for the fair-by-design ensemble models \(f^\star\) (Definition~\ref{def:fair_model}) built for the aforementioned data sets. %

\emph{Stability profiles} are our first visual diagnostic tool; they provide an overview of predictive volatility with respect to the entire fairness validation set from the perspective of a group of classifiers drawn from the chosen model family \(\mathcal{F}\) -- see Figure~\ref{fig:experiments:stability_profile}. %
They capture the consistency of individual predictions across models belonging to distinct performance bands \(\epsilon\) by grouping them in colour-coded pyramids. %
Each such stack informs us of the number -- quantity of horizontal segments -- and frequency -- width thereof -- of unique prediction vectors (i.e., different class assignments) across the entire fairness validation data set. %
For example, if the evaluation yields the following prediction vectors for a given performance band \(\epsilon\): [0, 0, 0, 0], [1, 1, 1, 1], [0, 0, 0, 0] and [0, 1, 1, 0], the corresponding pyramid is constructed of three segments whose widths are 2, 1 and 1. %
The most desirable, i.e., the most cross-model fair, shape of such a pyramid is one amassing all the models in a single segment at the bottom as this configuration indicates that every classifier offers the same set of predictions. %
The least desirable shape is a stack of segments whose width is 1, i.e., each model returns a different prediction vector. %

\emph{Fairness profiles} complement stability profiles by focusing on the behaviour of each model with respect to individual instances as shown in Figure~\ref{fig:experiments:fairness_profile}. %
They visualise how distinct classifiers -- one per row -- grouped together by predictive performance bands -- captured by colour-coding -- assign a (binary) decision -- differentiated by the saturation of each cell -- to every instance -- one per column -- in the designated validation data set. %
For example, the collection of 36 rows coloured in orange in Figure~\ref{fig:experiments:fairness_profile:credit_aproval} visually depicts the binary predictions -- light or dark shading -- output by 36 models whose accuracy is 93\% for 69 instances represented by individual columns. %
These plots shed light on the volatility of predictions across individuals, hence cross-model fairness of their classification. %
The stability and fairness profiles are linked through shared colouring of the predictive performance bands, for example, compare Figures~\ref{fig:experiments:stability_profile:credit_aproval} and~\ref{fig:experiments:fairness_profile:credit_aproval}. %
A fairness profile can be plotted \emph{faithfully}, thus accurately depicting the classification of every instance for all the models; %
alternatively, a \emph{summary} variant that aggregates distinct predictions across models within a performance band (i.e., a coloured segment) by sorting them for each instance (i.e., a column) separately may instead be preferred due to its improved readability -- we use this approach in Figure~\ref{fig:experiments:fairness_profile}. %
Ideally, each column ought to be in a single -- light or dark -- shade, which indicates a consistent, thus fair, prediction output by models in a single level or throughout different levels of predictive performance.%

\begin{figure}[b!]
    \centering
    \begin{subfigure}[t]{0.96\linewidth}
        \centering
        \includegraphics[width=1.\textwidth]{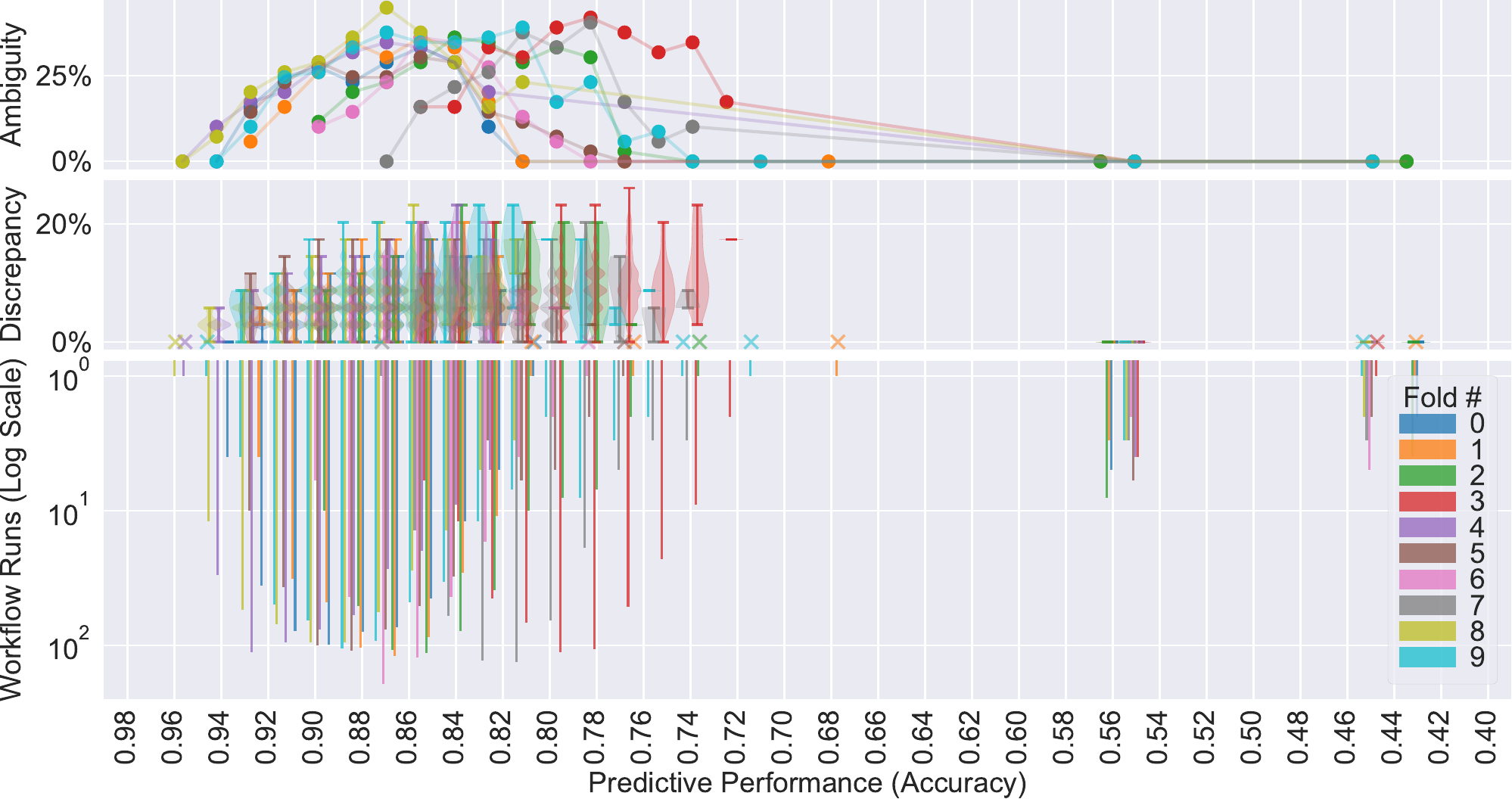}%
        \caption{Credit Approval.%
                 \label{fig:amb_dics:credit_approval}}
    \end{subfigure}
    \\[1em]%
    \begin{subfigure}[t]{0.96\linewidth}
        \centering
        \includegraphics[width=1.\textwidth]{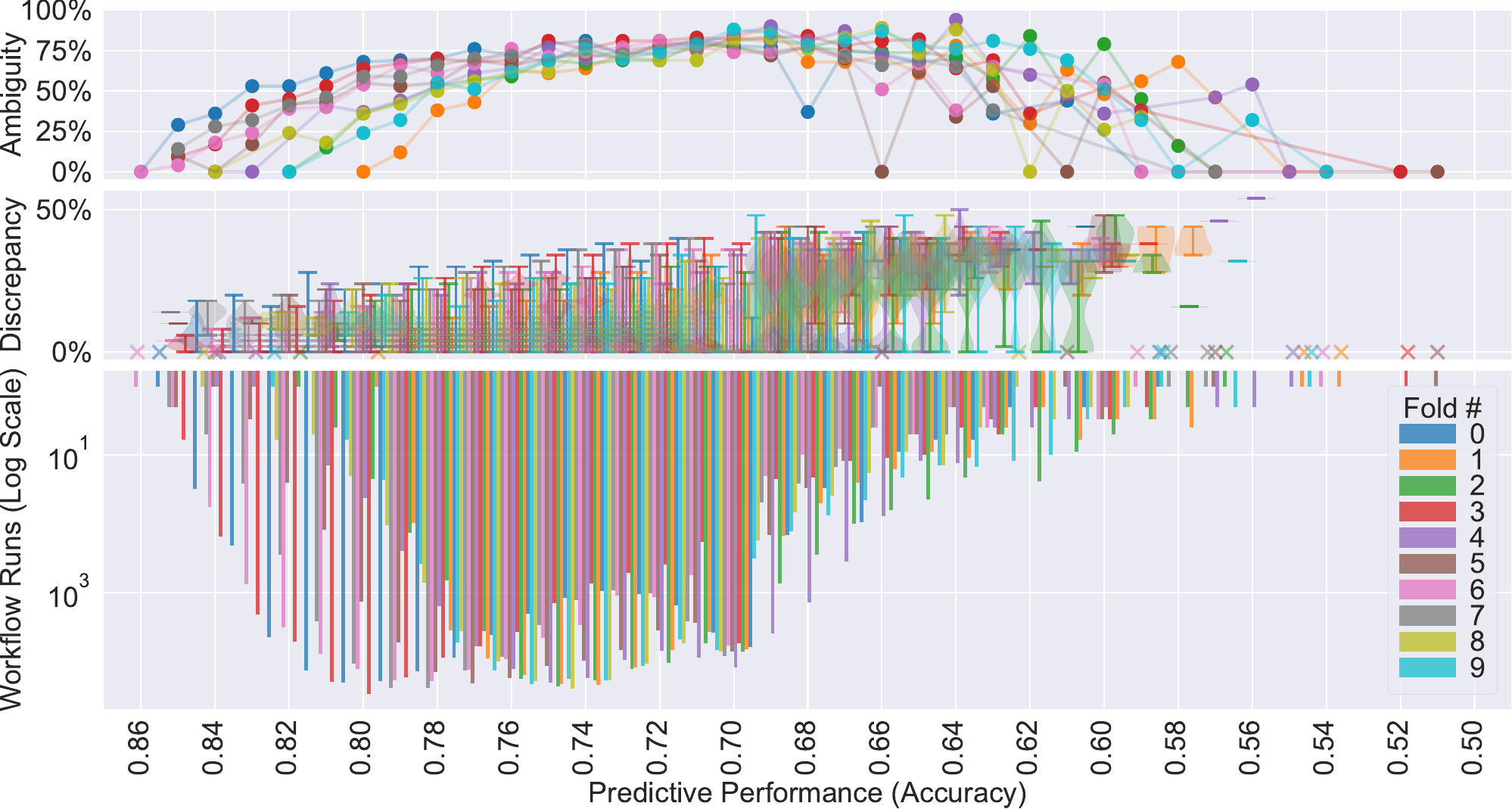}%
        \caption{German Credit.%
                 \label{fig:amb_dics:german_credit}}
    \end{subfigure}
\end{figure}
\begin{figure}\ContinuedFloat%
    \centering
    \begin{subfigure}[t]{0.96\linewidth}
        \centering
        \includegraphics[width=1.\textwidth]{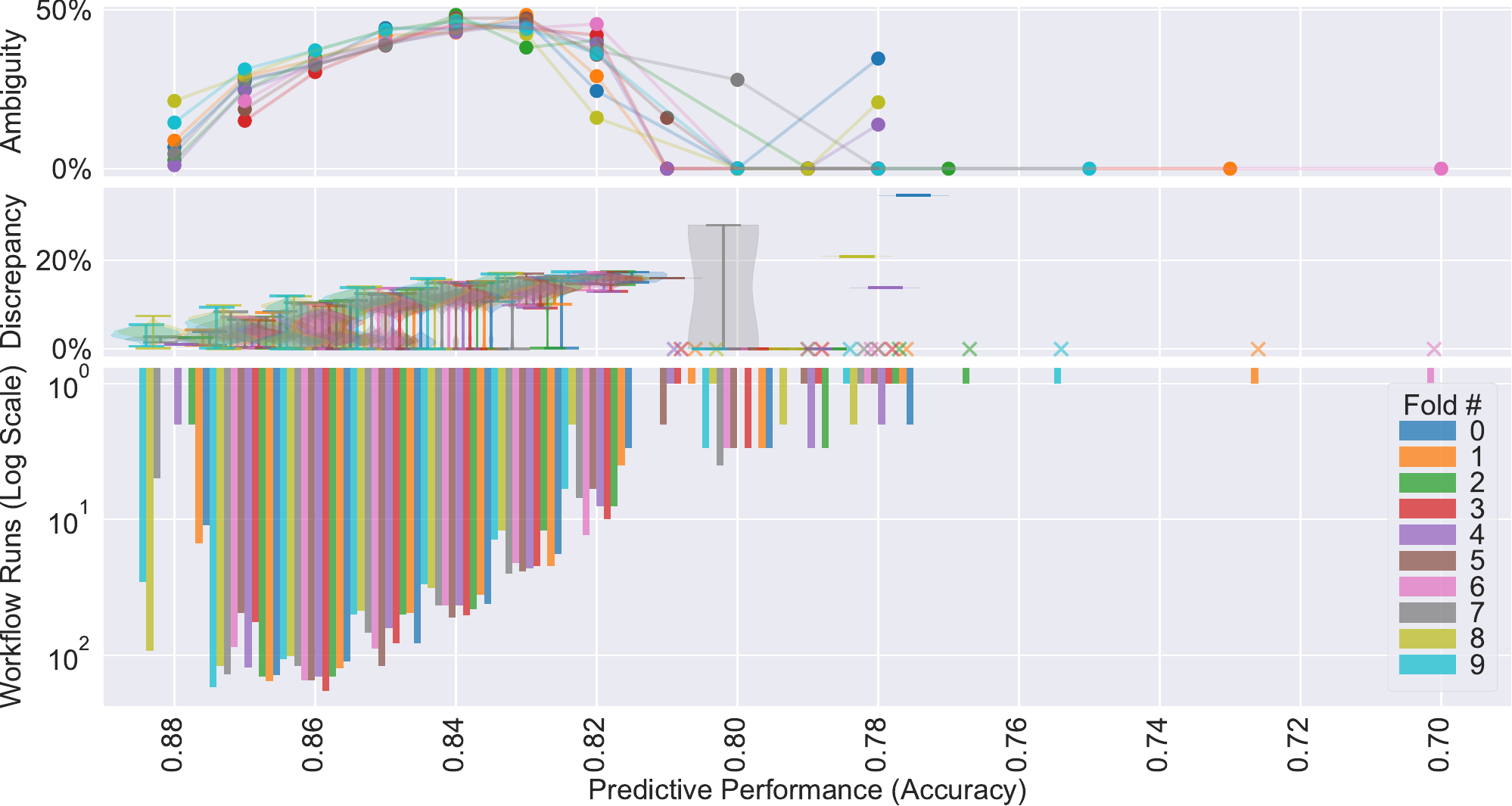}%
        \caption{Adult. Performance bands are relaxed -- \(\epsilon\simeq10^{-2}\) -- to improve readability.%
                 \label{fig:amb_dics:adult_2}}
    \end{subfigure}
    \caption{%
    Ambiguity (top), discrepancy (middle) and count of predictive models, i.e., workflow executions, (bottom, logarithmic scale) across performance bands for each data set. %
    Since each execution of an OpenML workflow is run on 10-fold cross validation, we have access to 10 distinct training--test data splits and their corresponding models; %
    we calculate these metrics separately for each fold-specific test data treating them as the fairness validation set to get a more detailed view of cross-model fairness (specifically, its stability). %
    Ambiguity calculates the (percentage) proportion of instances subject to unfairness in view of utility-based model multiplicity; %
    discrepancy measures the (percentage) proportion of instances for which predictions change between any two performance-equivalent models. %
    The (\subref{fig:amb_dics:credit_approval})~Credit Approval and (\subref{fig:amb_dics:german_credit})~German Credit data sets are displayed \emph{without} relaxing the performance bands. %
    The (\subref{fig:amb_dics:adult_2})~Adult data set is shown with the performance bands relaxed by rounding them to the 2\textsuperscript{nd} decimal place, i.e., \(\epsilon\simeq10^{-2}\), to improve readability of the plot (see Figure~\ref{fig:adult_amb} for ambiguity without rounding and with \(\epsilon\simeq10^{-3}\)). %
    The discrepancy for each performance band is based on up to 500 (randomly selected) workflow runs given the overwhelming number of their pairs for the full collection of models. %
    The \(\times\) symbol in the discrepancy plots indicates a single run of a predictive workflow, and the \(-\) marker is a flat violin plot where all workflows offer identical predictions.%
    \label{fig:amb_dics}}
\end{figure}

The remaining instrument in our toolkit -- shown in Figure~\ref{fig:amb_dics} -- is a visual depiction of two multiplicity metrics: ambiguity and discrepancy~\cite{marx2020predictive}. %
As explained earlier, each execution of an OpenML workflow is run on 10-fold cross validation, which yields 10 distinct pairs of training and test data sets together with the same number of the corresponding models. %
By calculating these metrics separately for each fold-specific test data (using its respective model), treating them as the fairness validation set, we get a more detailed view of cross-model fairness stability. %
\emph{Ambiguity} quantifies the (percentage) proportion of instances treated unfairly among models with a given level of predictive performance from a single family. %
\emph{Discrepancy} measures the (percentage) proportion of instances for which predictions change between any two models with a given level of predictive performance from a single family. %
We present the former as scatter plot-based trajectories, with flat shapes hovering near 0\% being optimal for cross-model fairness under predictive multiplicity -- they indicate consistent classification of individuals. %
We depict the latter as violin plots, showing the overall distribution as well as the minimum and maximum number of instances treated unfairly when switching between two arbitrary models with a given predictive performance level -- short violins capture more stable, hence cross-model fair, treatment of individuals. %
Both figures are accompanied by a bar plot illustrating (on a logarithmic scale) the number of distinct models, i.e., workflow executions, across all utility bands to help better assess the trustworthiness and reliability of these results. %

Before moving on to the analysis and discussion of our results, %
we inspect the influence of using non-zero tolerance \(\delta\) (Definition~\ref{def:multiplicity_delta}), i.e., relaxing the performance bands, on the quantity of unique utility levels (computed with accuracy). %
We investigate this phenomenon by rounding predictive performance \(\epsilon\) to the third (\(\epsilon\simeq10^{-3}\)) and second (\(\epsilon\simeq10^{-2}\)) decimal place, with the results reported in Table~\ref{tab:data_details}. %
While applying these tolerance values has no effect on Credit Approval and German Credit -- given that their performance validation sets have only up to 100 instances -- it affects Adult with its 4,885 data points. %
This observation suggests that for large validation sets we may need to relax the performance bands to get digestible and meaningful results -- as we will see later when analysing our findings -- however such an approach makes it easier to identify alternative models and challenge cross-model fairness. %
In general, this procedure is likely to degrade our notion of fairness as it combines predictions from across strict, i.e., non-relaxed, performance bands, creating a more diverse sample that possibly disagrees on a larger subset of data points. %

\begin{table}[t]
    \centering
    \caption{%
    Count of test instances and number of performance bands with different relaxation criteria for each selected data set. %
    Specifically, we study the quantity of unique utility levels when predictive performance \(\epsilon\) measured with accuracy is rounded to the 3\textsuperscript{rd} (\(\epsilon\simeq10^{-3}\)) and 2\textsuperscript{nd} (\(\epsilon\simeq10^{-2}\)) decimal place as well as without rounding (\(\epsilon\)). %
    The number of performance bands is likely to decrease when the utility measurements are relaxed, especially so for large validation sets.%
    \label{tab:data_details}}%
    \begin{tabular}{rrrrr}
        \toprule
        \multirow{2}{*}{Data Set} & \multirow{2}{*}{Test Points} & \multicolumn{3}{c}{Performance Bands} \\
        \cmidrule{3-5}%
                        &           & \makesamewidth[r]{\(\epsilon\simeq10^{-3}\)}{\(\epsilon\)} & \(\epsilon\simeq10^{-3}\) & \(\epsilon\simeq10^{-2}\) \\
        \midrule%
        Credit Approval & 69      & 12      & 12      & 12      \\
        German Credit   & 100     & 26      & 26      & 26      \\
        Adult           & 4,885   & 185     & 57      & 9       \\
        \bottomrule
    \end{tabular}
\end{table}

Equipped with the right insights and tools, we are in a position to assess cross-model fairness under utility-based predictive multiplicity for the selected data sets. %
Figures~\ref{fig:experiments:stability_profile} and \ref{fig:experiments:fairness_profile} show their \emph{stability} and \emph{fairness profiles}; %
Figure~\ref{fig:amb_dics} depicts the \emph{ambiguity} and \emph{discrepancy} metrics of model multiplicity. %
As explained earlier, the former two display results only for the first fold of each workflow execution, whereas the latter spans all of the folds. %
Given that our analysis produces a large number of (accuracy-based) performance bands, the stability and fairness profiles show only a collection of top-performing models. %

First, we focus on the stability profile of each selected data set shown in Figure~\ref{fig:experiments:stability_profile}. %
While for the Credit Approval and German Credit data sets (Figures~\ref{fig:experiments:stability_profile:credit_aproval} and \ref{fig:experiments:stability_profile:german_credit}) the high-utility classifiers appear cross-model fair -- the stack corresponding to the most accurate models (left-most, depicted in blue) has only one segment -- this ceases to hold for classifiers with lower accuracy. %
In contrast, the predictive workflow used for the Adult data set (Figure~\ref{fig:experiments:stability_profile:adult}) generates a large collection of models whose accuracy differs beyond the second decimal point -- multiple stacks composed of at most 3 segments of width 1 -- highlighting potential cross-model fairness issues. %
One contributing factor may be significantly larger predictive performance and fairness validation sets: 4,885 instances as compared to 69 and 100 for the other two data sets; %
however, as we proceed with our investigation we see that this size discrepancy is not entirely to blame.%

Next, we move on to the summary variant of the fairness profiles generated for the selected data sets -- shown in Figure~\ref{fig:experiments:fairness_profile} -- to paint a more accurate picture of their cross-model fairness. %
Recall that consistent shading within each column, i.e., for an individual data point, both within a single and across the colour-coded performance bands, conveys a stable, thus cross-model fair, prediction. %
A vertical bar spanning a single performance band that changes its saturation in the middle signifies the maximum disagreement between the models of this utility. %
While the profiles for the Credit Approval and German Credit data sets (Figures~\ref{fig:experiments:fairness_profile:credit_aproval} and \ref{fig:experiments:fairness_profile:german_credit}) appear relatively consistent, %
the one for Adult (Figure~\ref{fig:experiments:fairness_profile:adult}) is more jittery (note that for legibility reasons the plot only shows a subset of 250 instances out of 586 individuals who are treated unfairly, with the entire data set composed of 4,885 points).%

\begin{figure}%
    \begin{subfigure}[t]{.42\linewidth}%
        \centering
        \includegraphics[width=.77\textwidth]{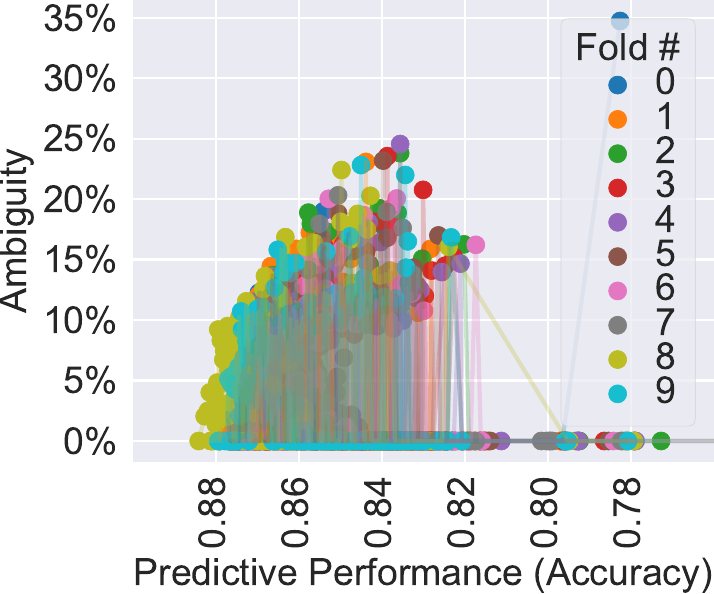}%
        \caption{Adult without rounding of performance bands.%
                 \label{fig:adult_amb:0}}
    \end{subfigure}
    \hspace{.04\textwidth}%
    \begin{subfigure}[t]{0.42\linewidth}%
        \centering
        \includegraphics[width=.77\textwidth]{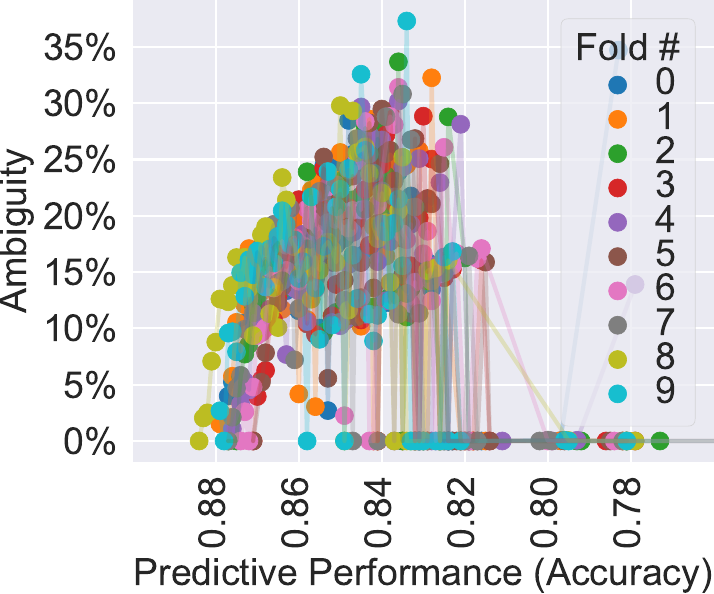}%
        \caption{%
        Adult with \(\epsilon\simeq10^{-3}\) rounding of performance bands.%
                 \label{fig:adult_amb:3}}
    \end{subfigure}
    \caption{%
    Ambiguity range for the Adult data set increases from (\subref{fig:adult_amb:0}) 0--25\% to (\subref{fig:adult_amb:3}) 0--35\% when predictive performance bands are rounded to the 3\textsuperscript{rd} decimal place. %
    The plots are truncated on the right side for readability; see Figure~\ref{fig:amb_dics:adult_2} for a reference of the full scale and comparison to rounding performance bands to the 2\textsuperscript{nd} decimal place, i.e., \(\epsilon\simeq10^{-2}\), which yields ambiguity range of 0--50\%.%
    \label{fig:adult_amb}}
\end{figure}

To further investigate cross-model fairness under predictive multiplicity we analyse ambiguity and discrepancy of our predictive workflows; %
the results are shown in Figure~\ref{fig:amb_dics} and corroborate the insights gathered thus far. %
Note that for the Credit Approval and German Credit data sets (Figures~\ref{fig:amb_dics:credit_approval} and \ref{fig:amb_dics:german_credit}) we work with strict model multiplicity \(\mathcal{F}_\epsilon\) (Definition~\ref{def:multiplicity}), whereas Adult (Figure~\ref{fig:amb_dics:adult_2}) is investigated with relaxed performance bands \(\mathcal{F}_{\epsilon\simeq\delta}\) (Definition~\ref{def:multiplicity_delta}), which in this case is achieved by rounding utility to the second decimal point (\(\epsilon\simeq10^{-2}\)). %
We apply this procedure since the stability (Figure~\ref{fig:experiments:stability_profile:adult}) and fairness (Figure~\ref{fig:experiments:fairness_profile:adult}) profiles for this particular data set suggest that ambiguity and discrepancy may otherwise be uninformative. %
Based on the former we can see an overwhelming number of unique prediction vectors -- visible as stacked segments of width 1 -- in each colour-coded performance band, signifying cross-model unfair treatment of many instances; the latter plot paints a similar picture. %
We validate these observations by calculating the number of performance bands as well as computing ambiguity without and with rounding to the third (\(\epsilon\simeq10^{-3}\)) and second (\(\epsilon\simeq10^{-2}\)) decimal points -- refer to Table~\ref{tab:data_details} and compare Figures~\ref{fig:amb_dics:adult_2} and \ref{fig:adult_amb}.%

\begin{figure}%
    \centering
    \begin{subfigure}[t]{0.32\textwidth}
        \centering
        \includegraphics[width=1.\linewidth]{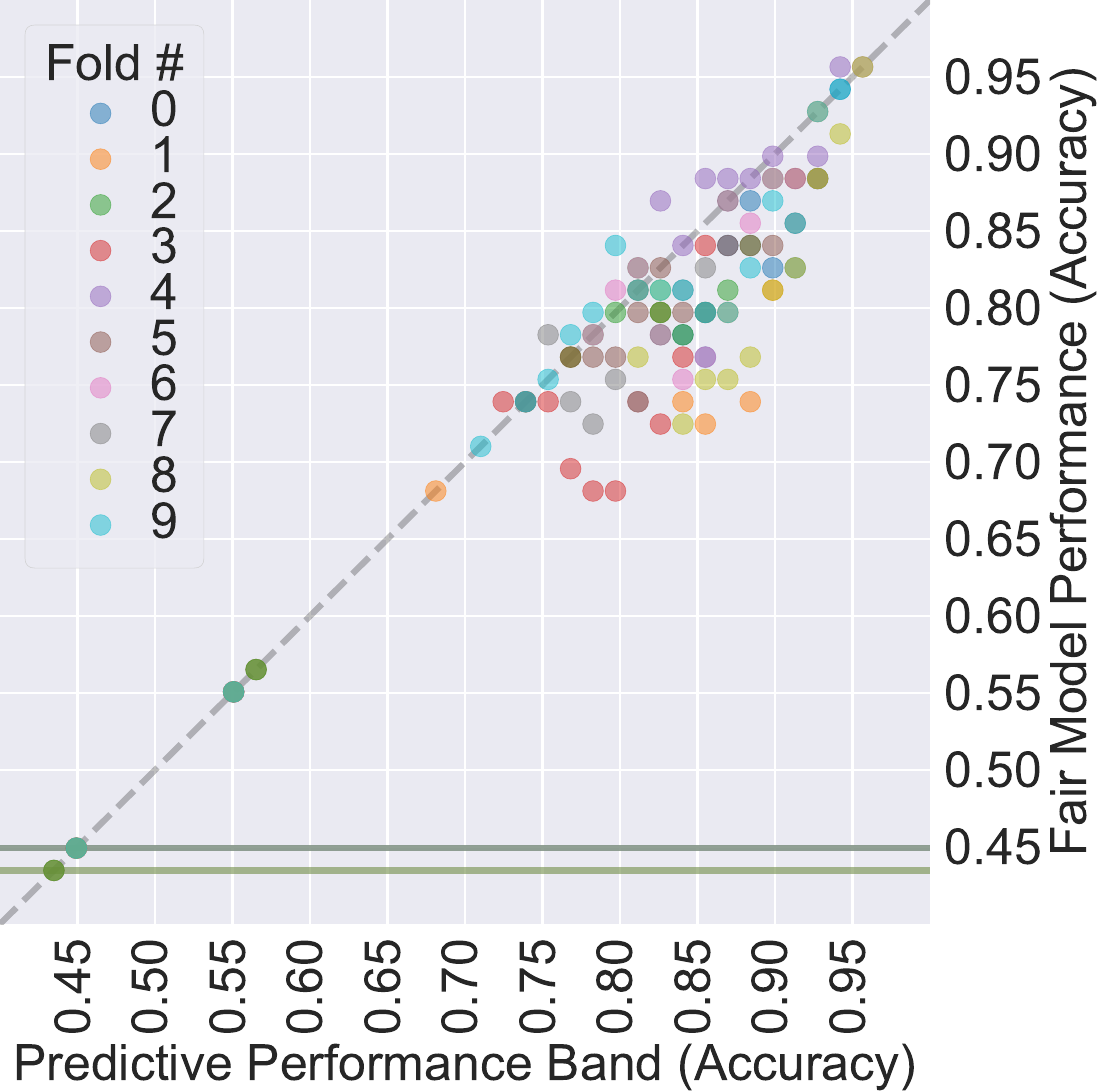}%
        \caption{Credit Approval.%
                 \label{fig:perfect_model:credit_approval}}
    \end{subfigure}
    \hspace{.015\textwidth}%
    \begin{subfigure}[t]{0.32\textwidth}
        \centering
        \includegraphics[width=1.\linewidth]{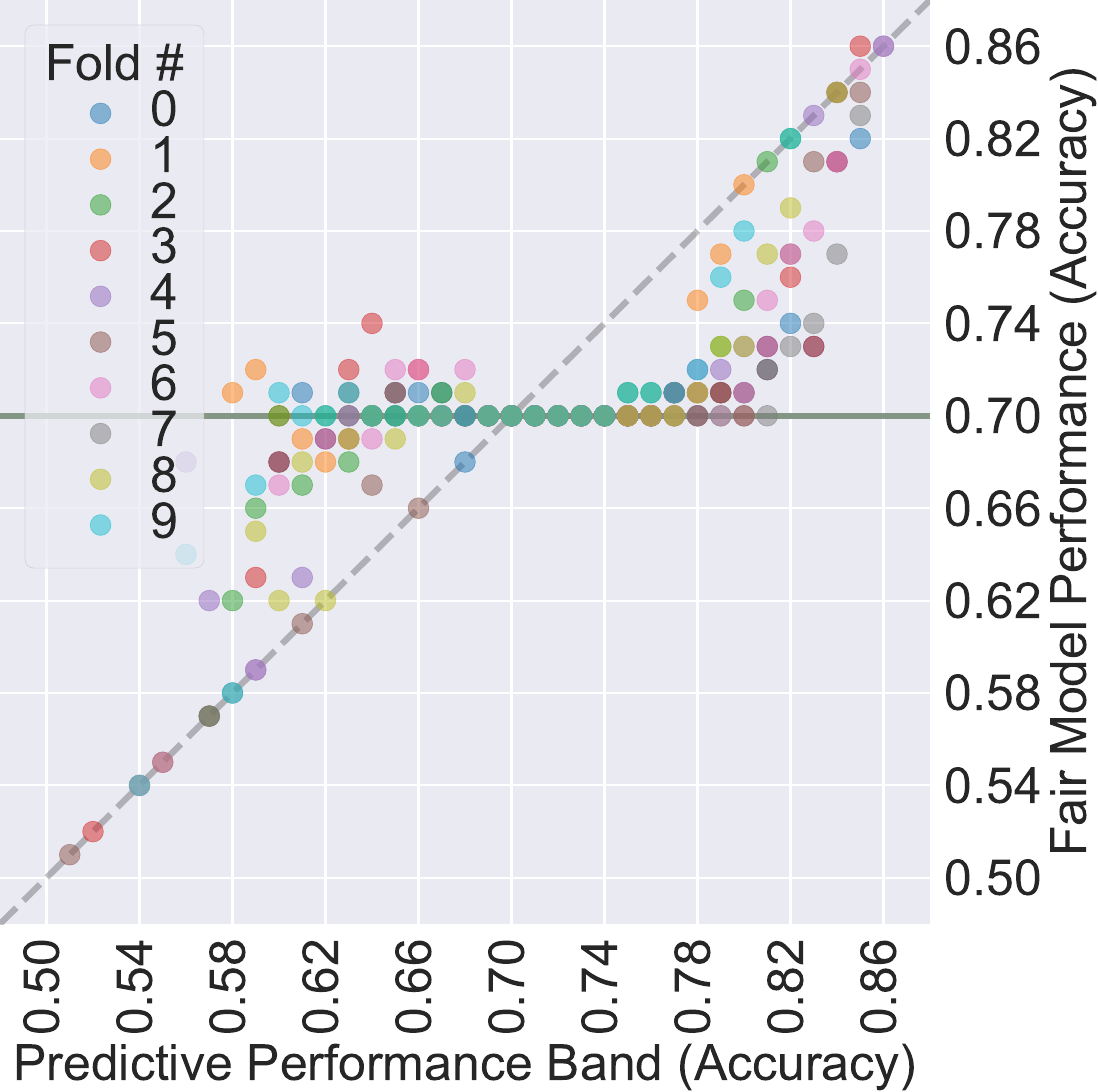}%
        \caption{German Credit.%
                 \label{fig:perfect_model:german_credit}}
    \end{subfigure}
    \\[1em]%
    \begin{subfigure}[t]{0.32\textwidth}
        \centering
        \includegraphics[width=1.\linewidth]{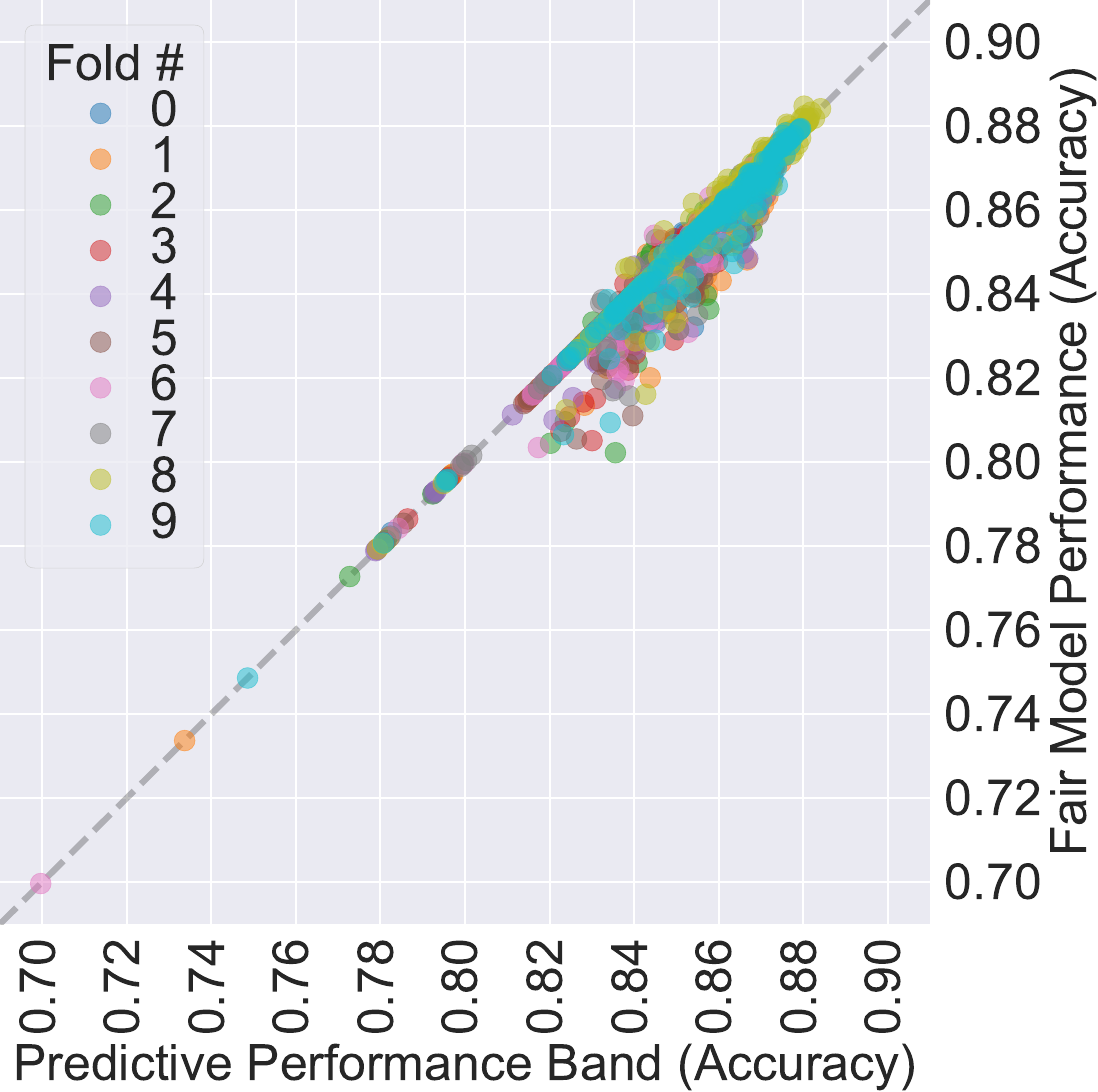}%
        \caption{Adult.%
                 \label{fig:perfect_model:adult_0}}
    \end{subfigure}
    \hspace{.015\textwidth}%
    \begin{subfigure}[t]{0.32\textwidth}
        \centering
        \includegraphics[width=1.\linewidth]{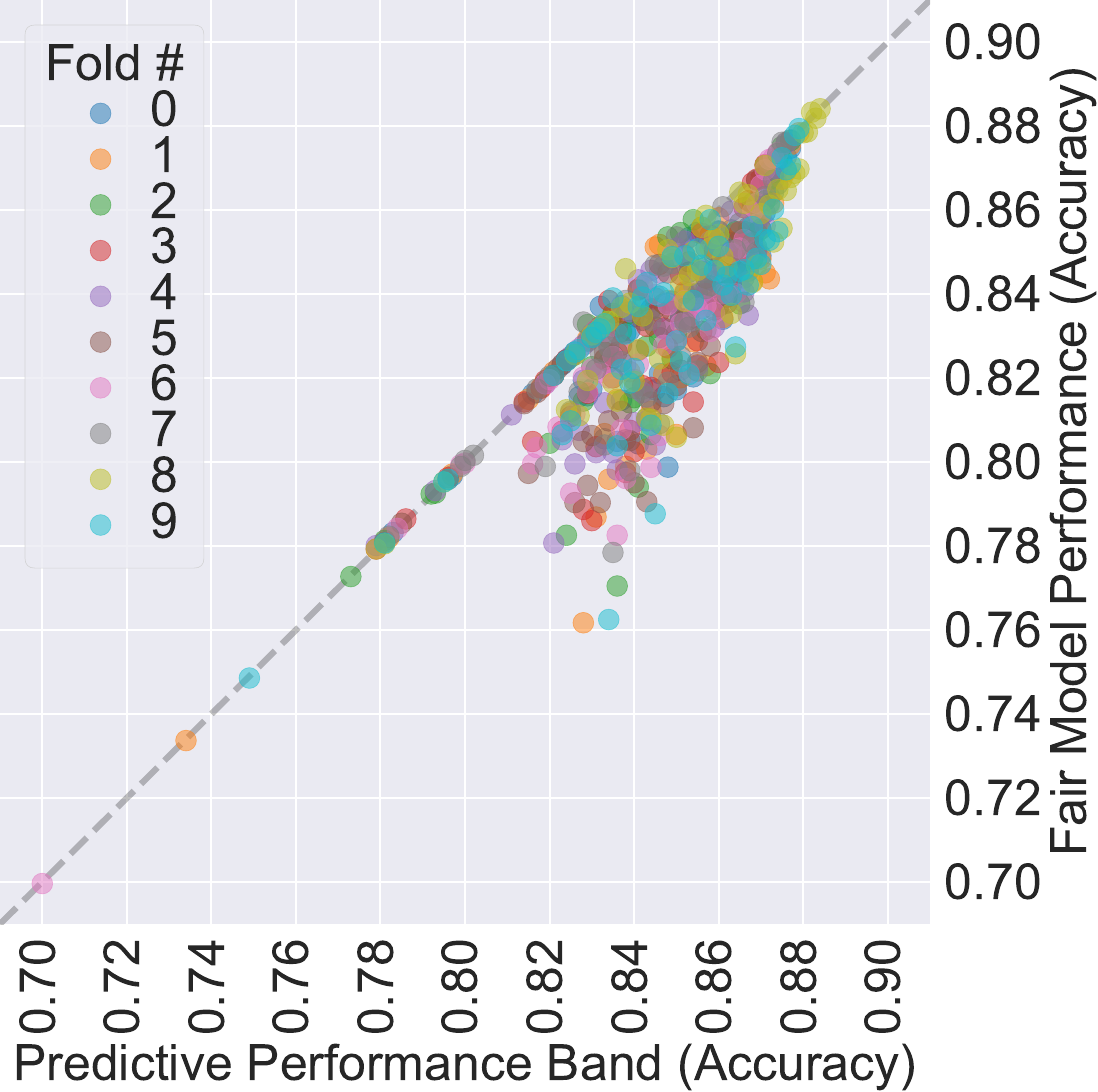}%
        \caption{Adult with \(\epsilon\simeq10^{-3}\).%
                 \label{fig:perfect_model:adult_3}}
    \end{subfigure}
    \hspace{.015\textwidth}%
    \begin{subfigure}[t]{0.32\textwidth}
        \centering
        \includegraphics[width=1.\linewidth]{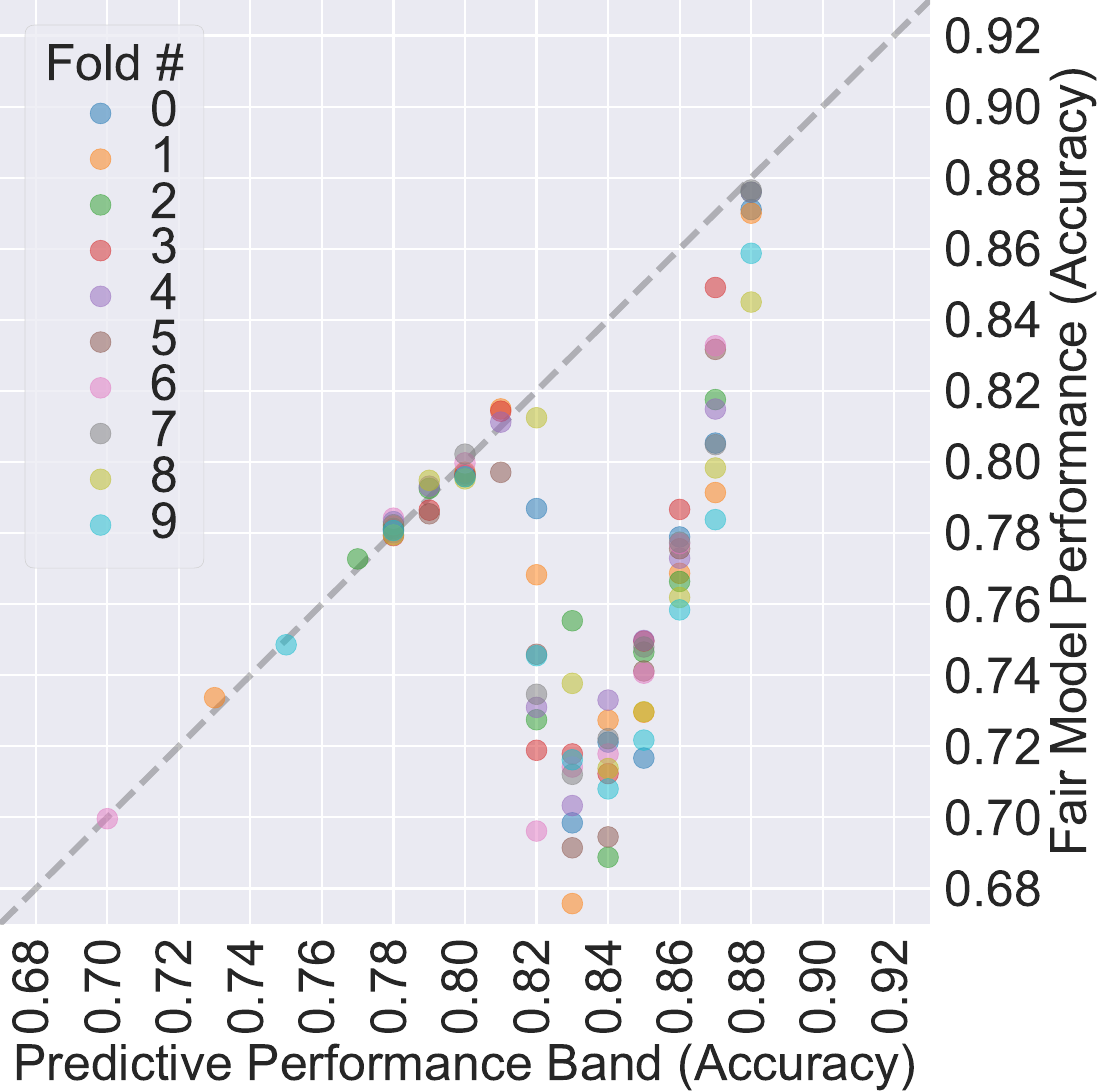}%
        \caption{Adult with \(\epsilon\simeq10^{-2}\).%
                 \label{fig:perfect_model:adult_2}}
    \end{subfigure}
    \caption{%
    Predictive performance (accuracy) of the fair-by-design ensemble models \(f^\star\) (y-axis) compared to (\subref{fig:perfect_model:credit_approval}, \subref{fig:perfect_model:german_credit} \& \subref{fig:perfect_model:adult_0}) strict \(\epsilon\) and (\subref{fig:perfect_model:adult_3} \& \subref{fig:perfect_model:adult_2}) relaxed \({\epsilon\simeq\delta}\) performance bands (x-axis) discovered for individual models \(f \in \mathcal{F}_\epsilon\) and \(f \in \mathcal{F}_{\epsilon\simeq\delta}\) respectively. %
    For consistency, only Adult is investigated under relaxed utility. %
    The diagonal line marks unchanged performance, with points above indicating an improvement and points below a decrease in the accuracy of the individually fair models built upon their respective collections of fixed-utility classifiers. %
    The horizontal lines, drawn separately for each fold, indicate the performance achieved by always predicting the preferred class; most of them overlap in Panels~(\subref{fig:perfect_model:credit_approval}) and (\subref{fig:perfect_model:german_credit}), and they are not visible (24\%) in Panels~(\subref{fig:perfect_model:adult_0}), (\subref{fig:perfect_model:adult_3}) and (\subref{fig:perfect_model:adult_2}). %
    Overall, we see up to 15\% drop in accuracy for the fair model, but this pattern is broken for data with highly imbalanced classes as shown by Panel~(\subref{fig:perfect_model:german_credit}).%
    \label{fig:perfect_model}}
\end{figure}

For all the data sets, ambiguity raises quite sharply as we move away from an optimal model, i.e., the top performance band, reaching between 40 and 90 per cent at its peak, which draws attention to a surprisingly high degree of cross-model unfairness for these classifiers. %
Discrepancy shows that while in many cases it is possible to switch between two (performance-wise) equivalent models without affecting individual predictions, a non-negligible proportion of instances -- captured by tall violin plots -- is likely to be treated differently. %
More generally, the predictive workflows used with the Credit Approval and German Credit data sets seem to behave consistently without the need of relaxing performance bands, whereas the one used with Adult only provides discernible patterns when utility is rounded to the third (\(\epsilon\simeq10^{-3}\)) or second (\(\epsilon\simeq10^{-2}\)) decimal point. %
We suspect that in the latter case the culprits are volatility and high expressiveness of the underlying predictive model -- decision tree-based AdaBoost -- however this intuition cannot be confirmed without further analysis. %
Notably, relaxing performance bands merges distinct levels of utility, thereby introducing more disagreements on the level of individual predictions. %
This phenomenon can be observed in the ambiguity range being stretched from 0--25\% without rounding the predictive performance to 0--35\% and 0--50\% when it is rounded to the third and second decimal places respectively (see Figures~\ref{fig:amb_dics:adult_2} and \ref{fig:adult_amb} for reference).%

Finally, we analyse the utility -- measured with accuracy -- of the fair-by-design ensemble models \(f^\star\) (Definition~\ref{def:fair_model}) constructed for every performance band of each selected data set. %
In addition to strict model multiplicity \(\mathcal{F}_\epsilon\) (Definition~\ref{def:multiplicity}), we investigate relaxed performance bands \(\mathcal{F}_{\epsilon\simeq\delta}\) (Definition~\ref{def:multiplicity_delta}) exclusively for Adult, rounding accuracy to the third (\(\epsilon\simeq10^{-3}\)) and second (\(\epsilon\simeq10^{-2}\)) decimal points. %
The results -- presented in Figure~\ref{fig:perfect_model} -- show a nearly universal drop in predictive performance when dealing with \(f^\star\). %
An interesting exception is German Credit -- Figure~\ref{fig:perfect_model:german_credit} -- for which the proportion of the favourable class is 70\% (based on the ground truth labels). %
This class imbalance improves the accuracy of individually fair models for performance bands (x-axis) below 70\%, whereas after this mark the utility drops as expected. %

The insights provided by our comprehensive inspection toolkit lead to a few important observations. %
When a classifier is sub-optimal in terms of its predictive performance -- which we may not necessarily know at the time -- its cross-model unfairness is especially concerning. %
In our experiments this can be seen in the stability profiles (Figure~\ref{fig:experiments:stability_profile}), where the height and width of the pyramids rapidly increase as we move towards lower utility bands; %
the same behaviour is reflected in the sharp raise in ambiguity as we move away from an optimal model (Figure~\ref{fig:amb_dics}). %
Plotting the latter metric across \emph{all} the utility bands allows us to pick up another interesting aspect of fairness in view of model multiplicity: incremental improvements of predictive performance may lack measurable influence on cross-model fairness (especially for sub-optimal models). %
This phenomenon is captured by the initial rate of change in ambiguity and the wide span of the worst (highest) value of this metric across numerous performance bands. %
While specific instances affected by cross-model unfairness may change as we improve the predictive performance of a model, metrics measuring this property may remain relatively unchanged until we cross a certain utility threshold -- a phenomenon that may be misleading to model developers. %

Another observation is the prevalence of cross-model unfairness for large fairness validation sets, which can be seen across our inspection toolkit for the Adult data set. %
While this is not necessarily a universal characteristic, such a behaviour highlights the importance of considering the difficulty of classifying a particular data set given the capabilities and expressiveness of the selected family of models. %
More specifically, we should inspect whether certain individuals are more prone than others to be subject to cross-model unfairness either within a subset of performance bands or across all of them -- discrepancy plots (Figure~\ref{fig:amb_dics}) communicate this phenomenon on a high level and fairness profiles (Figure~\ref{fig:experiments:fairness_profile}) allow to zoom in and see a more fine-grained perspective. %
Identifying the affected data points and generalising them into representative cohorts can help to devise decision heuristic to combat cross-model unfairness; %
their development can additionally account for domain-specific properties such as the anticipated class imbalance and the selected evaluation metric to deliver better predictive performance than a na{\"i}ve operationalisation of our notion of cross-model fairness, such as through the fair-by-design ensemble model. %
The following section offers a more theoretical discussion of cross-model fairness and outlines its broader implications. %

\section{Towards Cross-model Fairness of Individual Predictions\label{sec:multiplicity}}%

Our experiments show the prevalence and severity of cross-model unfairness in view of utility-based predictive multiplicity for real-life machine learning workflows, which prompts us to explore this problem on a more fundamental level. %
Such an investigation is especially important given that the na\"ive operationalisation of the proposed notion of fairness brings about various challenges, in particular a possible drop in predictive performance that curtails the usefulness of the underlying data-driven modelling effort. %
This phenomenon is acutely visible for the fair-by-design ensemble model given by Definition~\ref{def:fair_model} whose example is constructed in Figure~\ref{fig:linera_model_multiplicity_1_merged}, where \(f^\star\) misclassifies two instances whereas the base models \(f \in \mathcal{F}_\epsilon\) make just one mistake each. %
Among others, we therefore should pay special attention to the composition of the selected %
performance and fairness validation sets since they are respectively a key to model equivalence and identification of unfair behaviour. %
In particular, their representativeness, density and relative distribution should be carefully considered and potentially expanded or augmented over time. %

In addition to the modelled data space, we ought to scrutinise the inherent characteristics of the classifiers themselves. %
Overall stability of the model family, which may depend on the stochasticity or greediness of the corresponding training procedure, as well as proximity of instances to class boundaries -- for example, conveyed by prediction uncertainty and model (over- or under-) confidence -- and behaviour of a classifier in sparse data regions all play a role in the volatility of automated decision-making. %
The inherent expressiveness and flexibility of the chosen model, and more broadly any data modelling workflow that can be built upon it by incorporating new steps such as data pre-processing or feature engineering, can also be problematic. %
In particular, it may lead to \emph{pathological} instantiations of the fair-by-design ensemble model where \emph{every} individual is offered the most favourable outcome, which we discuss in the next paragraph. %
All of these observations create opportunities (predominantly on technical grounds) for individuals adversely affected by an automated decision to easily challenge cross-model fairness of this process in view of utility-based predictive multiplicity. %

One conceivable attack on our notion of cross-model fairness can come from \emph{underspecification} of the selected family of classifiers \(\mathcal{F}\) or any predictive workflow built around it. %
Expressive classifiers may be flexible enough to single out individual instances and assign an arbitrary prediction to them, for example, due to their inherent complexity or parameterisation scope. %
Such freedom can adversely influence cross-model fairness by permitting any two instances to swap predictions -- regardless of their placement in the data space -- while maintaining the desired level of performance \(\epsilon\), i.e., operating within the designated utility-based model multiplicity class \(\mathcal{F}_\epsilon\). %
This can be easily achieved for classifiers with considerable parameterisation scope such as deep neural networks, but it is also relevant to simpler models, e.g., \(k\)-nearest neighbours, when they are misconfigured as shown in Figure~\ref{fig:neighbours_model_multiplicity_1}.%

\newlength{\mytextwidth}
\setlength{\mytextwidth}{\textwidth}

\begin{figure}[t]
    \centering
    \begin{subfigure}[t]{0.485\linewidth}%
        \centering
        \includegraphics[width=.23\mytextwidth]{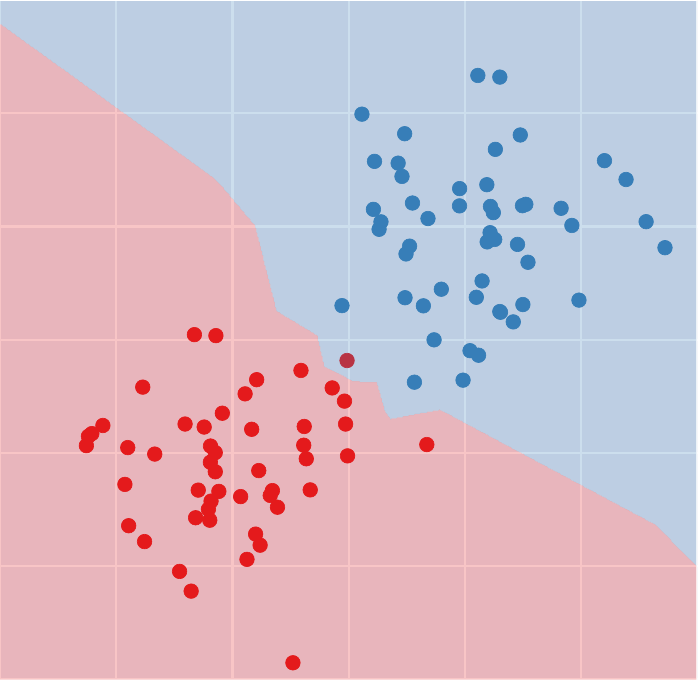}%
        \hspace{1em}%
        \includegraphics[width=.23\mytextwidth]{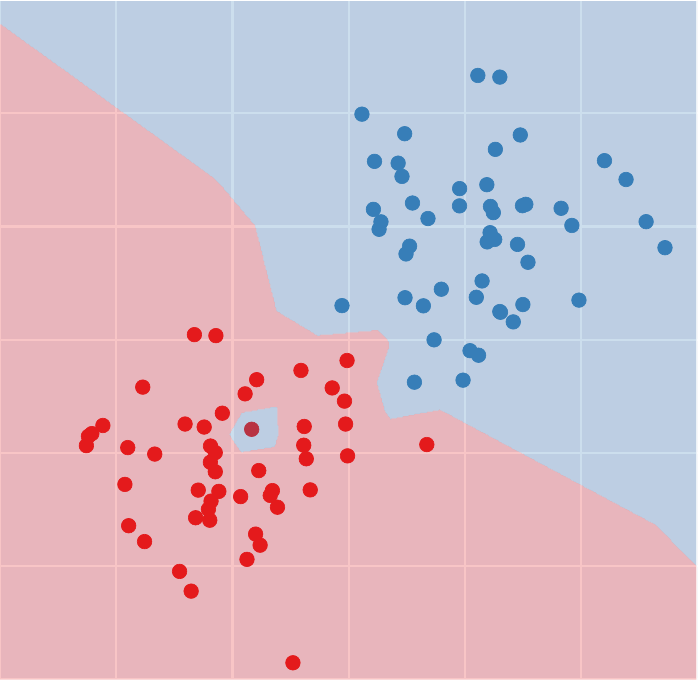}%
        \caption{One red instance is misclassified.\label{fig:neighbours_model_multiplicity_1:2}}%
    \end{subfigure}
    \hfill%
    \begin{subfigure}[t]{0.485\textwidth}
        \centering
        \includegraphics[width=.23\mytextwidth]{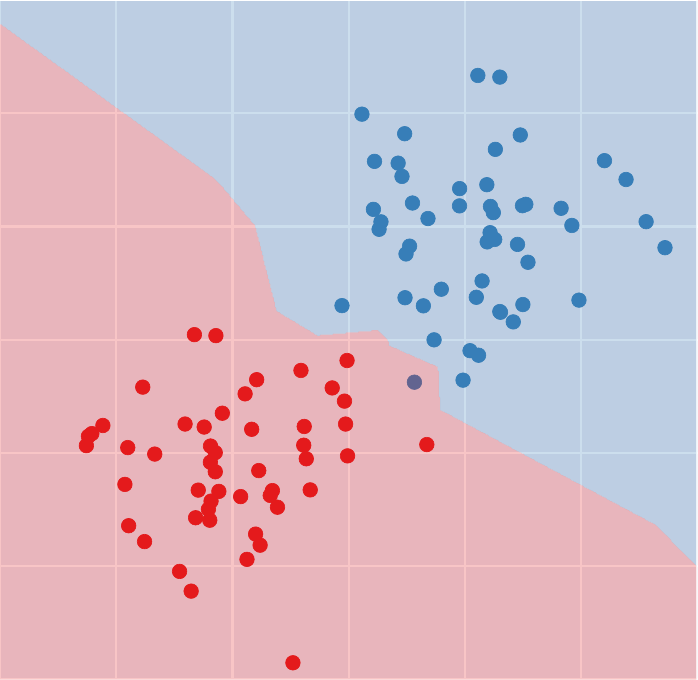}%
        \hspace{1em}%
        \includegraphics[width=.23\mytextwidth]{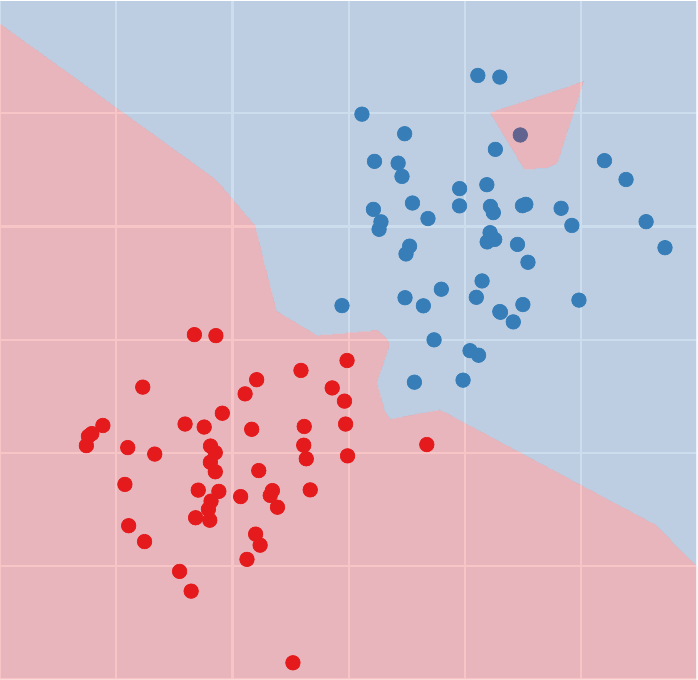}%
        \caption{One blue instance is misclassified.\label{fig:neighbours_model_multiplicity_1:1}}%
    \end{subfigure}
    \caption{Model multiplicity with one error (99\% accuracy) for a \(k\)-nearest neighbours classifier with \(k = 1\). %
    Expressive models can be flexible enough to allow for an arbitrary classification of every instance, e.g., by manipulating training data. %
    To mitigate such scenarios appropriate safeguards need to be established through constraints on the model form and its parameterisation (e.g., \(k \geq 42\)) or regularisation being required as part of a predictive workflow built on top of a selected model family \(\mathcal{F}\).%
    \label{fig:neighbours_model_multiplicity_1}}
\end{figure}

Flexibility of a model \(f\) from a given family \(\mathcal{F}\) can be defined through a proxy such as complexity \(\Omega(f)\) or Vapnik--Chervonenkis dimension~\cite{vapnik1971uniform}, and imposed as an additional constraint. %
The precise specification of such a metric may be unique to each model family; for example, the number of non-zero parameters for linear models, the highest coefficient degree for polynomial classifiers and the depth, width or number of instances per leaf for decision trees. %
More broadly, expressiveness, hence flexibility, of models and workflows built upon them may be controlled by ensuring diversity of training data, fixing a lower bound on confidence of each decision, abstaining from predictions, restricting model parameterisation, enforcing regularisation such as pruning for trees and LASSO for linear models, or limiting overfitting via alternative mechanisms. %
Otherwise, with a budget of errors given by the required level of predictive performance, an excessive number of people could abuse this notion and claim unfair treatment.%

The aforementioned strategies tasked with constraining the model space can be complemented by (use case-specific) operational requirements. %
For example, instead of a single performance metric employed to determine model equivalence, their hierarchy can be implemented, measuring accuracy first, and followed by precision and recall to address any ties. %
More broadly, such an approach could be applied directly to the confusion matrix of a classifier by sequentially imposing restrictions on its individual entries. %
Arguably, one could optimise exclusively for \emph{recall} and \emph{specificity} to maximise the number of favourable decisions; however, %
any such course of action is just a stopgap as it is likely to lack a solid foundation rooted in the domain-specific aspects of the modelled problem. %
A more meaningful heuristic should therefore rely on well-defined properties of the utilised model~\cite{breiman2001statistical} -- for example, monotonicity of a particular feature with respect to the prediction -- which strategy is consistent with some definitions of AI and ML interpretability~\cite{sokol2021explainability}. %
A particularly promising solution that marries technical properties and relevant aspects of the underlying data domain is imposing constraints on the model form~\cite{rudin2019stop,sokol2023reasonable}. %
Such an approach ensures a strong notion of (ante-hoc) interpretability and address the problems posed by cross-model fairness when faced with predictive multiplicity, which are especially pronounced for flexible model classes. %
This solution is particularly promising given that both of these desiderata are of paramount importance for high stakes domains, and constrained model forms can help us to fulfil both of them. %

Lacking a (close to) \emph{unique} classifier -- obtained by imposing various desiderata to narrow down the scope of the chosen family of predictive models, thus making any alternative difficult or impossible to find -- we may need to instead rely on the fair-by-design classifier \(f^\star\) outlined by Definition~\ref{def:fair_model}. %
Treating each individual with the best available model in this fashion may however degrade the overall predictive performance for the task at hand, making the fair application of data-driven decisions on a par with less effective classifiers to begin with. %
For example, this phenomenon can be observed for the fair-by-design model \(f^\star\) shown in Figure~\ref{fig:linera_model_multiplicity_1_merged}. %
Notably, any model family \(\mathcal{F}_\epsilon\) subject to \emph{population-} or \emph{validation-based} multiplicity, i.e., with meaningful alternatives, is bound to suffer from disputable spaces. %
Depending on the density of data in these regions and their size, the fair-by-design model \(f^\star\) will offer \emph{no worse recall} but \emph{no better specificity} than any individual model \(f \in \mathcal{F}_\epsilon\) as stated by Proposition~\ref{col:fair_model}. %
This change in predictive performance is especially prominent (reaching its limit) when the chosen family of models \(\mathcal{F}\) is expressive enough to single out individual data points, in which case the cross-model fair-by-design classifier \(f^\star\) will assign the most favourable outcome to every instance -- 0\% specificity and 100\% recall -- as given by Proposition~\ref{col:fair_model_expressive}. %
While undesirable, this behaviour is a manifestation of an inherent \emph{trade-off} between cross-model (individual) fairness and utility -- since enforcing the former implicitly introduces modelling constraints that curtail the latter -- which is not unique to our notion of fairness~\cite{friedler2021possibility} and which needs to be carefully managed, especially in high stakes domains. %

\begin{proposition}\label{col:fair_model}%
    \textbf{Specificity} \(m_s\) of the cross-model individually fair-by-design classifier \(f^\star\) under utility-based predictive multiplicity \(\mathcal{F}_\epsilon\) is no better than that of any other classifier \(f \in \mathcal{F}_\epsilon\):%
    \[
    \forall \; f \in \mathcal{F}_\epsilon \;\; m_s \left( f^\star(\mathcal{X}), \mathcal{Y} \right) \leq m_s \left( f(\mathcal{X}), \mathcal{Y} \right) \text{,}%
    \]
    where \((\mathcal{X}, \mathcal{Y})\) is the labelled data space. %
    Additionally, \textbf{recall} \(m_r\) of \(f^\star\) is no worse than that of any \(f \in \mathcal{F}_\epsilon\):%
    \[
    \forall \; f \in \mathcal{F}_\epsilon \;\; m_r \left( f^\star(\mathcal{X}), \mathcal{Y} \right) \geq m_r \left( f(\mathcal{X}), \mathcal{Y} \right) \text{.}%
    \]
\end{proposition}

\begin{proposition}\label{col:fair_model_expressive}%
    When a family of classifiers \(\mathcal{F}_\epsilon\) under utility-based predictive multiplicity is expressive enough to assign a selected class \(c \in \mathcal{Y}\) to any data point \(x \in \mathcal{X}\), i.e.,%
    \[
    \forall \; x \in \mathcal{X} \; \exists \; f \in \mathcal{F}_\epsilon \;\; \text{s.t.} \;\; f(x) = c \text{,}
    \]
    the corresponding cross-model individually fair-by-design classifier \(f^\star\) achieves 0\% \textbf{specificity} (\(m_s\)) and 100\% \textbf{recall} (\(m_r\)):%
    \[
    m_s \left( f^\star(\mathcal{X}), \mathcal{Y} \right) = 0\% \qquad m_r \left( f^\star(\mathcal{X}), \mathcal{Y} \right) = 100\% \text{,}%
    \]
    in which case every point is assigned the most favourable outcome.%
\end{proposition}

The results above, specifically Proposition~\ref{col:fair_model}, can be interpreted geometrically as expanding the space predicted with the more favourable outcome (red in our case) to the entirety of disputable regions -- see Figure~\ref{fig:linera_model_multiplicity_1_merged} for an example. %
It follows directly from observing individual predictions under \(f^\star\):%
\begin{itemize}
    \item Each data point from the preferred class (red) predicted as such (true positive) will not be affected, but misclassified positive instances (false negatives) may be corrected by \(f^\star\) -- a possible improvement in \textbf{recall}.%
    \item Each data point from the undesirable class (blue) predicted as such (true negative) may be (mis)classified as positive, but misclassified negative instances (false positives) will not be affected by \(f^\star\) -- a possible drop in \textbf{specificity}.%
\end{itemize}
Furthermore, the fair-by-design classifier \(f^\star\) for models \(f \in \mathcal{F}_\epsilon\) from a family that is flexible enough to predict any individual with an arbitrary class will assign the preferred output to all data, from which Proposition~\ref{col:fair_model_expressive} follows. %
These observations alone should encourage the developers of AI and ML systems to meaningfully constrain the model form and strive for high utility computed on a comprehensive and representative validation set to minimise the scope of any disputable regions. %

Given the benefits of finding a classifier \(f \in \mathcal{F}_\epsilon\) that is favourable for a specific individual, the concept of cross-model fairness under utility-based predictive multiplicity can be framed as an adversarial challenge. %
In such a setting people negatively affected by an automated decision can confront the owner of the underlying predictive model by building an equivalent classifier (performance-wise) that offers them the desired outcome instead. %
These ``adversaries'' attempt to identify a predictor \(f^\prime\) from within the employed family of models \(\mathcal{F}_\epsilon\) -- ensuring that it complies with all of the restrictions imposed by \(\mathcal{F}_\epsilon\), including the predetermined level of predictive performance \(\epsilon\) -- that assigns a selected individual the most favourable class. %
The party responsible for building and deploying the challenged model, on the other hand, ought to minimise the possible number of such claims by reducing the size of any disputable regions. %
This tug-of-war is bound to have a positive effect on the predictive model in question by iteratively improving its accountability, robustness and overall quality. %
Two possible realisations of this process are through \emph{collective action} or by establishing dedicated \emph{oversight bodies}. %
The former can accumulate and track cross-model unfairness though strategies such as crowdsourcing, the success stories of which are the Ad Observatory~\cite{adobservatory} and Gender Shades~\cite{buolamwini2018gender} projects; %
the latter can be given unrestricted access to all the required components and attempt to build alternative models in pursuit of uncovering adversely affected individuals. %
Reporting requirements are another possibility that found success in different areas of AI and ML, e.g., Model Cards~\cite{mitchell2019model}. %

Facilitating this back-and-forth process may require releasing (a subset of) relevant training data as well as performance and fairness validation sets (in addition to the specification of the utilised model family \(\mathcal{F}\)) -- either exclusively to oversight bodies or to the broader public -- which may be problematic due to inherent trade secrets. %
While distributing training data cannot be easily sidestepped, predictive performance of a model may be assessed without access to evaluation data as the classifier itself can instead be submitted for testing, akin to how ML competitions are run. %
Notably, having a comprehensive performance validation set that faithfully represents all of the individual cases is advantageous for the model creators as it restricts the number of admissible classifiers. %
Lastly, publishing a fairness validation set may also be beneficial to the owner of the model under investigation, encouraging a broader community to identify unfair predictions -- and, more generally, disputable spaces -- as well as engendering trust in the deployed model itself. %
This decoupling of the two -- predictive performance validation data and fairness validation set -- may therefore be desirable, especially that the latter \emph{does not need to be labelled} since as it stands we are only interested in disparate treatment of these instances across equivalent classifiers \(f \in \mathcal{F}_\epsilon\) when assessing cross-model individual fairness.%

Satisfying the strictest definition of cross-model (individual) fairness may be impossible, impractical or undesirable; for example, because of the aforementioned utility trade-off or a need to disclose sensitive information, which challenges arise for many other notions of algorithmic fairness~\cite{kleinberg2017inherent,small2022robust}. %
Our discussion thus far has been predominantly focused on the technical aspects of cross-model fairness and its enforcement strategies, but given the limited help that such approaches appear to offer we can implement high-level \emph{modelling requirements} and employ model \emph{selection heuristics} as well as apply well-motivated \emph{prediction aggregation strategies}~\citep{black2022model}. %
The former two are intended to abate model underspecification and tame the degree of multiplicity, whereas the latter is meant to harness the diverse range of (possibly opposing) predictions instead of ignoring them, thus provide a transparent justification of the (cross-model) fairness of the final decision, %
with the realisation of these processes reflecting the level of impact of the underlying application domain. %
By formalising meta-rules for model specification, construction and selection, we can guide the modelling efforts by imposing additional (individual or group) fairness, complexity, robustness or transparency constraints and operational limitations, e.g., training data availability, compute power budget and execution time restrictions, as well as adherence to ethical norms and societal priorities. %
Given that, in principle, there is an infinite number of performance-equivalent predictors, we should be able to identify a collection of models that best capture our values and fulfil the chosen desiderata. %

While limiting the scope of multiplicity, such approaches are unlikely to guarantee cross-model fairness, achieving which may still require the use of the fair-by-design ensemble model (Definition~\ref{def:fair_model}). %
Depending on the use case, however, such a strict prediction aggregation strategy -- whose implications and trade-offs we discussed at length -- may not always be appropriate or desirable~\citep{black2022model}. %
Instead of outputting the most favourable prediction we can rely on a \emph{majority vote}, select a model \emph{at random} from the multiplicity set, or draw the final decision according to the \emph{distribution of individual predictions} given by performance-equivalent models (with a similar randomisation practice used to enforce group fairness~\citep{small2023equalised}). %
The suitability of each aggregation method needs to be determined and justified on a case-by-case basis, which process will offer the grounds for avoiding arbitrary decisions as well as motivate the principles and reasoning behind (a particular variant of) cross-model fairness. %
Such alternatives can prove particularly useful when dealing with allocation of limited resources (for which applicants are deemed equally eligible) or highly imbalanced predictions output by the multiplicity set, in which cases randomising the final decision may be the fairest solution; %
e.g., when 5 positive decisions can be awarded but the fair-by-design ensemble model offers such an outcome to 10 individuals, or when only 1 out of 500 performance-equivalent models provides the desired output. %

Just as predictive robustness is an important desideratum of a data-driven model for AI and ML engineers, cross-model fairness ought to be of concern to people being affected by automated decision-making. %
Guaranteeing this property engenders trust in model-makers who in such a scenario have to make the best effort to offer each individual an undisputable outcome (from a technical perspective). %
Additional reassurance can be had by giving people the benefit of the doubt whenever they fall into a disputable region; this can either be the most favourable prediction when operationalising our fair-by-design ensemble model or a bespoke decision heuristic chosen based on the context and application domain. %
Notably, such a scenario is a specific instance of favouring the user in the face of uncertainty, quantified here through predictive multiplicity. %
This approach is in line with augmenting classification with a reject or abstain option to account for scenarios where a model lacks confidence to predict an instance. %

The definition of cross-model fairness relies heavily on the selected family of data-driven classifiers and any modelling restrictions imposed upon it. %
This has wide-reaching implications: the fairness of the selected model is not only contingent on its own predictive behaviour but also on the behaviour of other classifiers within the same utility-based multiplicity regime, regardless of whether the model-maker is aware of their existence. %
While this is a relatively strong requirement that imposes most of the burden on the creators of data-driven models, it is intended to instil a high level of confidence in people subject to algorithmic decision-making, who are relatively powerless in comparison to the organisations collecting the data and building the models. %
It incentivises the model makers to impose constraints on the model form as well as continuously monitor and revise their modelling pipelines -- which approach is especially important for high stakes domains~\cite{rudin2019stop} -- but also to forgo relying purely on data-driven methods (which tend to exhibit high stochasticity) and ad hoc selection of the final model since such practices are likely to broaden the scope of the underlying model family. %
As each individual prediction may necessitate an explanation or justification depending on its operational context, or even require facilitating an opportunity for contestation and recourse, cross-model fairness frees people from having to pursue the aspect of this process linked to classification non-uniqueness. %
Additionally, the difficulty in satisfying this notion of fairness is likely to promote designing human-in-the-loop predictive systems or operating them in the decision support mode instead, thus encouraging responsible adoption of data-driven automation. %
The challenges of implementing cross-model fairness -- such as having to precisely define the admissible family of models and searching through it to identify any disputable spaces -- may nonetheless only warrant all this effort when the stakes are high enough. %

\section{Conclusion and Future Work\label{sec:conclusion_future}}%

In this paper we formalised the notion of cross-model fairness that arises due to utility-based model multiplicity -- a scenario in which a collection of classifiers assigns different labels to selected data points despite being considered equivalent in terms of predictive power. %
Such a situation is pertinent to distinct types of models, diverse parameterisation of the same model, or classifiers that result from multiple training runs of a given model when this procedure is, for example, greedy or stochastic. %
Additionally, high-dimensional and sparse data may contribute to this phenomenon given the curse of dimensionality (i.e., everything being far away from each other in high-dimensional spaces). %
In particular, we built these concepts around a user-specified \emph{family of predictive models}, taking into consideration its \emph{expressiveness}, which in the extreme may lead to awarding the most favourable prediction to every individual. %
We then defined \emph{two meaningful notions of model multiplicity}: one determined by the entire data space and another measured with respect to a designated \emph{fairness validation data set}. %
We generalised the former into \emph{disputable spaces} -- regions where at least two models disagree on a prediction -- which may not necessarily be identified even with a comprehensive fairness validation set. %
Next, we showed how to combine (performance-wise) equivalent models to present each individual with the best possible decision. %

After establishing these foundations, we reported results of a comprehensive, large-scale, empirical study of cross-model fairness on three real-life data sets popular in the literature using top-performing classification workflows published in the OpenML repository. %
Such an experimental setup captures a realistic diversity of models that arises naturally in the machine learning workflow instead of being artificially constructed specifically to study model multiplicity, which ensures credibility of our findings. %
To help identify cross-model unfairness stemming from predictive multiplicity we adapted two relevant metrics -- \emph{discrepancy} and \emph{ambiguity} -- and introduced a bespoke visualisation toolkit. %
Our findings highlight the prevalence and severity of this phenomenon, illustrating the importance of considering utility-based model multiplicity as a new dimension of algorithmic fairness. %

We also demonstrated that employing the fair-by-design ensemble model, which grants each individual the most favourable outcome, may adversely affect the overall predictive power of the classification task at hand, especially for highly expressive models. %
As an alternative, we discussed imposing restrictions on the model space to limit the number of viable alternatives, arguing for enforcing (operational) constraints that are meaningful to each individual modelling problem; e.g., prediction monotonicity with respect to a chosen attribute, which is also recognised as a strategy for introducing (ante-hoc) interpretability whereby predictions are aligned with human values and can be easily justified. %
This approach can be complemented with non-technical modelling requirements and model selection heuristics as well as prediction aggregation strategies other than the one utilised by the fair-by-design ensemble model. %
Otherwise, being able to determine disputable spaces -- ``grey areas'' or ``edge cases'' from a classification standpoint -- can allow to abstain from making a decision or engage a human expert in the process, thus partially alleviate the prevalence of cross-model unfair predictions. %

In future work, we will investigate numerical metrics and analytical tools to assess cross-model unfairness in a broader context, extending the notion beyond crisp binary classification. %
For example, we will study some of the techniques discussed briefly in Section~\ref{sec:related_work}. %
We will also look into strategies for integrating ground truth labels into calculation of cross-model fairness to capture the quality of each prediction. %
Moreover, we will look into developing methods to derive bounds on the most and least favourable treatment of each individual from within a given data set for a selection of predictive models. %
This should facilitate systematic identification of disputable and stable regions, possibly leading to a new optimisation objective -- minimise the former or maximise the latter -- for training cross-model fair, robust and effective predictors. %
We will also study the lower limit of utility for the fair-by-design classifier that always uses the most beneficial model from their collection determined by a given level of predictive performance. %
Finally, we will explore diverse prediction aggregation strategies for performance-equivalent models and investigate their suitability for different use cases, leading to alternative, less strict, notions of cross-model fairness. %

\renewcommand{\acksname}{Acknowledgements}
\begin{acks}
This research was supported by %
the Estonian Research Council (projects number PUT1458 and PRG1604); %
the Dora Plus programme, sponsored by the European Regional Development Fund and Estonian government; %
the ARC Centre of Excellence for Automated Decision-Making and Society, funded by the Australian Government through the Australian Research Council (project number CE200100005); and %
the Hasler Foundation (grant number 23082). %
\end{acks}

\bibliographystyle{ACM-Reference-Format}
\bibliography{acm}

\end{document}